\documentclass{article} 
\pdfoutput=1
\usepackage{iclr2026_conference,times}

\usepackage{amsmath,amsfonts,bm}

\def\eqref#1{equation~\ref{#1}}

\def\1{\bm{1}}

\def\vs{{\bm{s}}}

\DeclareMathAlphabet{\mathsfit}{\encodingdefault}{\sfdefault}{m}{sl}
\SetMathAlphabet{\mathsfit}{bold}{\encodingdefault}{\sfdefault}{bx}{n}

\definecolor{citecolor}{rgb}{0.052,0.11,0.508}
\definecolor{linkcolor}{rgb}{0.9, 0.06, 0.06}
\usepackage[colorlinks=true,citecolor=citecolor, linkcolor=linkcolor]{hyperref}
\usepackage{url}
\usepackage{booktabs}       
\usepackage{amsfonts}       
\usepackage{nicefrac}       
\usepackage{microtype}      
\usepackage{xcolor}         
\usepackage{amsmath}

\usepackage{multirow}     
\usepackage{graphicx}     
\usepackage{cleveref}
\usepackage{xspace}
\usepackage{amssymb}
\usepackage{subcaption}
\usepackage{adjustbox}
\usepackage{floatrow}
\usepackage{graphicx}
\usepackage{url}            
\usepackage{booktabs}       
\usepackage{nicefrac}       
\usepackage{microtype}      
\usepackage{wrapfig}

\usepackage{multirow}
\usepackage{tablefootnote}

\usepackage{color}
\usepackage{colortbl}
\usepackage{xcolor}

\definecolor{blgrey}{rgb}{0.6,0.6,0.6}
\definecolor{bblue}{rgb}{0.855,0.933,0.98}
\definecolor{dblue}{HTML}{5297D6}
\definecolor{gainred}{rgb}{0.1,0.5,0.3}
\definecolor{citecolor}{HTML}{0071BC}
\definecolor{linkcolor}{HTML}{ED1C24}

\newcommand{\graycell}[1]{\textcolor{gray!60}{#1}}
\newcommand{\ablanum}[1]{\textcolor{dblue}{#1}}         

\makeatletter
\DeclareRobustCommand\onedot{\futurelet\@let@token\@onedot}
\def\@onedot{\ifx\@let@token.\else.\null\fi\xspace}

\def\eg{\emph{e.g}\onedot}

 \def\vs{\emph{vs}\onedot}

\makeatother

\usepackage{pifont}

\usepackage{enumitem}

\def\eqref#1{(\ref{#1})}

\def\y{\mathbf{y}}
\def\encoder{{\mathcal{E}}}
\def\decoder{{\mathcal{D}}}
\def\quantizer{{\mathcal{Q}}}
\def\codebook{{\mathcal{C}}}
\def\hatcodebook{{\hat{\mathcal{C}}}}
\def\c{{\mathbf{c}}}
\def\ze{{\mathbf{z}_e}}
\def\zq{{\mathbf{z}_q}}
\def\d{{d}}
\def\x{{\mathbf{x}}}
\def\haty{\mathbf{\hat{y}}}
\def\loss{\mathcal{L}}
\def\T{\mathcal{T}}
\def\zeset{{\mathcal{P}_z}}
\def\zqset{{\mathcal{H}_z}}
\def\cset{{\mathcal{C}_z}}
\def\hatcset{{\hat{\mathcal{C}_z}}}
\def\monitor{VQBridge}
\def\model{FVQ}

\newtheorem{observation}{Observation}

\title{Scalable Training for Vector-Quantized Networks with 100\% Codebook Utilization}

\author{%
  Yifan Chang$^{1,4}$ \quad\quad
  Jie Qin$^{2}$ \quad\quad
  Limeng Qiao$^{2}$ \quad\quad
  Xiaofeng Wang$^{3}$ \\
  \textbf{Zheng Zhu$^{3}$} \quad\quad\quad\space
  \textbf{Lin Ma$^{2}$} \quad\quad
  \textbf{Xingang Wang$^{1,5}$} \\
  $^{1}$CASIA \quad\quad $^{2}$Meituan \quad\quad $^{3}$GigaAI \quad\quad $^{4}$UCAS \\
  $^{5}$Luoyang Institute for Robot and Intelligent Equipment
}

\iclrfinalcopy 
\arxivcopy

\begin{document}

\maketitle

\begin{figure}[htb]
\ificlrfinal
    \vspace{-10mm}
\fi
\begin{center}
    \includegraphics[width=1.0\textwidth]{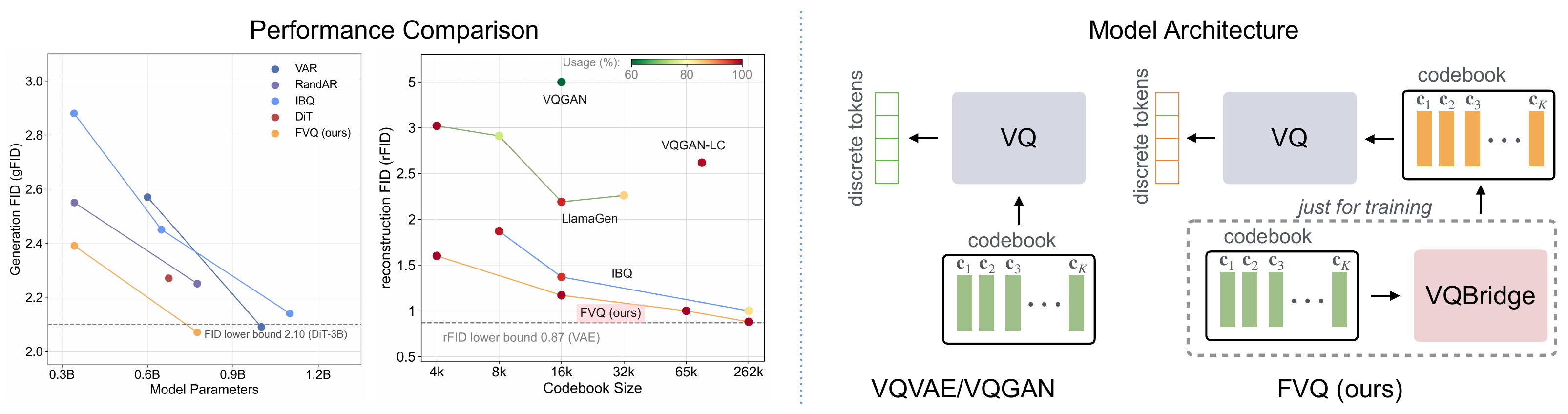}
\end{center}
\vspace{-4pt}
\caption{\small
\textbf{Performance comparison and model architecture}. 
\textbf{Left}: Performance comparison showing generation (FID) and reconstruction (rFID) quality across different methods. \textbf{Right}: Architectural comparison between baseline VQVAE/VQGAN and the proposed FVQ model, with VQBridge component for training.
\vspace{-4pt}
}
\label{fig:first}
\end{figure}

\begin{abstract}
Vector quantization (VQ) is a key component in discrete tokenizers for image generation, but its training is often unstable due to straight-through estimation bias, one-step-behind updates, and sparse codebook gradients, which lead to suboptimal reconstruction performance and low codebook usage. 
In this work, we analyze these fundamental challenges and provide a simple yet effective solution. 
To maintain high codebook usage in VQ networks (VQN) during learning annealing and codebook size expansion, we propose \monitor{}, a robust, scalable, and efficient projector based on the map function method. \monitor{} optimizes code vectors through a compress–process–recover pipeline, enabling stable and effective codebook training. 
By combining \monitor{} with learning annealing, our VQN achieves full (100\%) codebook usage across diverse codebook configurations, which we refer to as \model{} (FullVQ).
Through extensive experiments, we demonstrate that \model{} is effective, scalable, and generalizable: it attains 100\% codebook usage even with a 262k-codebook, achieves state-of-the-art reconstruction performance, consistently improves with larger codebooks, higher vector channels, or longer training, and remains effective across different VQ variants. Moreover, when integrated with LlamaGen, \model{} significantly enhances image generation performance, surpassing visual autoregressive models (VAR) by 0.5 and diffusion models (DiT) by 0.2 rFID, highlighting the importance of high-quality tokenizers for strong autoregressive image generation.
\ificlrfinal
\url{https://github.com/yfChang-cv/FVQ}
\fi
\end{abstract}    
\section{Introduction}
\label{sec:intro}

    The scalability of autoregressive models has been well demonstrated in large language models~\citep{gpt1, gpt3, llama1, llama3, qwen}. Inspired by this, pioneering works~\citep{vqgan, vit-vqgan, dalle, var} have explored the application of autoregressive models in visual generation tasks. To enhance scalability and reduce computational costs, a common approach~\citep{vqvae, vqvae2} involves introducing discrete tokenizers that employ vector quantization (VQ) to convert continuous images into discrete tokens. Autoregressive models are trained on these tokens in a self-supervised strategy to predict the next tokens. Generally, the reconstruction capability of the tokenizer directly determines the upper bound of the generative model's performance~\citep{vqgan, magvit2}. Therefore, the design of discrete tokenizers plays a crucial role in autoregressive visual generation.

    However, due to the information loss introduced by vector quantization, the performance of discrete tokenizers has long been limited compared with continuous tokenizers~\citep{vae}. The intuitive insight is that enlarging the codebook could reduce the information loss during quantization. Nevertheless, prior work~\citep{vit-vqgan, llamagen, unitok} has shown that simply increasing the codebook size or code dimension often leads to codebook collapse, which causes a severe reduction in codebook usage and consequently undermines the generative capability of downstream models~\citep{straightening, open-magvit2}. Despite many efforts to mitigate this issue, when scaling to larger settings (\eg, codebook size of 262k and code dimension of 256)~\citep{ibq}, full codebook usage remains unattainable and the performance is still unsatisfactory.

    To address this issue, we revisit VQ from fundamental mechanism and three fundamental challenges~\citep{straightening}, together with empirical observations that shed light on resolving them. VQ relies on the straight-through estimator (STE)~\citep{ste} to resolve the non-differentiability problem, but this introduces three additional challenges: straight-through estimation bias, one-step-behind update, and sparse codebook gradients. The first two challenges cause misalignment in optimization across the network, leading to instability, while sparse codebook gradients prevent the codebook from being sufficiently trained. In practice, only a small subset of code vectors are updated, which is the primary cause of collapse.

    Surprisingly, based on our observations, these challenges can be effectively mitigated using only learning annealing and a sufficient powerful projector that jointly optimizes the codebook. Prior works \citep{vqganlc, simvq} employ a linear layer as the projector for joint codebook optimization, showing some improvement. However, our further experiments reveal that a linear projector alone is fragile: it makes the network highly sensitive to the learning rate and insufficiently capable when scaling to larger codebooks. To enable robust and scalable VQ training, we propose a robust, scalable and efficient projector, called \monitor{}, as shown in Figure~\ref{fig:first}. \monitor{} adopts a compress–process–recover pipeline: it first compresses the set of code vectors into a smaller number, then models interactions via ViT blocks~\citep{vit}, and finally recovers the vectors to the original codebook size and dimensionality. This design allows VQ to rapidly achieve and sustain 100\% codebook usage for arbitrary codebook configurations, enabling effective optimization of VQ.

    Using \monitor{} combined with learning annealing, we conduct extensive experiments to validate its effectiveness, scalability, and generalization. On the ImageNet benchmark~\citep{imagenet}, \model{} consistently achieves 100\% codebook usage across all configurations, including a 262k-codebook, while substantially enhancing the reconstruction performance of discrete tokenizers (rFID = 0.88), without introducing additional parameters or inference cost. \model{} exhibits strong scalability, with performance consistently improving as the codebook size, vector channel, and number of training epochs increase. Furthermore, the approach remains effective when applied to multi-code representation VQ~\citep{rqvae, var}, demonstrating robust generalization. Besides, \model{} leads to substantial improvements in generation performance: for instance, LlamaGen-XL improves FID from 3.39 to 2.07, outperforming comparable visual autoregressive models VAR~\citep{var} (FID=2.57) and diffusion models DiT~\citep{dit} (FID=2.27). These results highlight that even a vanilla autoregressive framework can achieve strong generation capabilities when the tokenizer exhibits high-quality reconstruction.

    In summary, our contributions include:

    1. We analyze the fundamental challenges arising from vector quantization and provide effective observations, leading to a simple yet scalable training framework (\model{}).
    
    2. Through extensive experiments, we show that \model{} is effective, scalable, and generalizable. Specifically, it achieves 100\% codebook usage across all configurations, attains state-of-the-art reconstruction performance, and consistently improves as the codebook size, vector channel, and number of training epochs increase, remains effective across different types of VQ.
    
    3. We demonstrate that \model{} substantially enhances the generative capability of autoregressive models. Without any bells and whistles, a vanilla autoregressive framework equipped with \model{} outperforms advanced visual autoregressive models (VAR) and diffusion models (DiT).

\section{Backgroud}
\label{backgroud}

\begin{figure}[ht]
\begin{center}
    \includegraphics[width=1\linewidth]{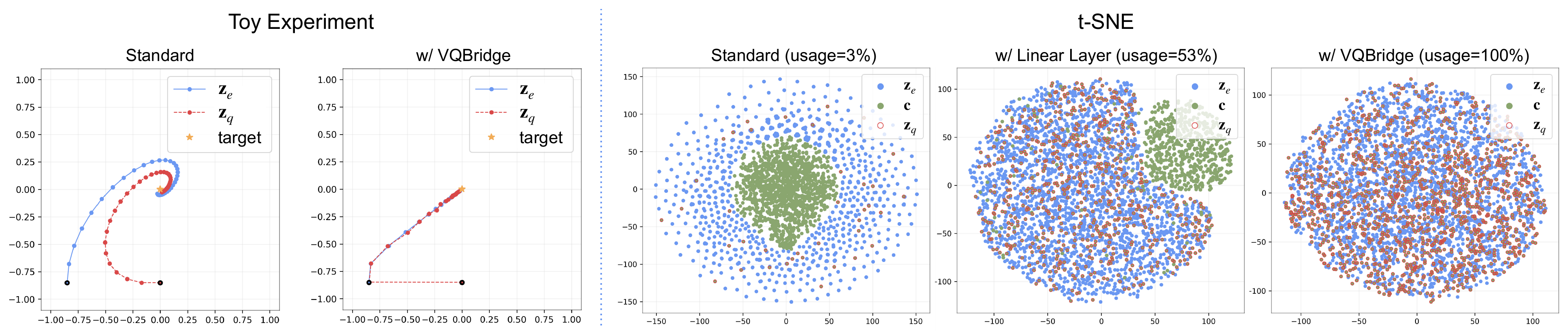}
\end{center}
\caption{\small
\textbf{Toy experiment and t-SNE visualization results.} \textbf{Left}: A dynamic vector quantization toy experiment where the target is a fixed optimization objective and the model is optimized with SGD. Using the standard method with only STE leads to a spiral trajectory, while \monitor{} produces a piecewise linear trajectory. With \monitor{}, $\zq$ closely follows $\ze$, reducing their distance already in the early stages of training. \textbf{Right}: Due to sparse gradients from the codebook, the standard method achieves only 3\% codebook usage. Adding a linear layer for joint optimization improves the usage but remains insufficient, whereas \monitor{} achieves 100\% usage.
}
\label{fig:toy}
\end{figure}

\textbf{Vector Quantized Network.} A vector quantized network (VQN) contains Encoder~$\encoder(\cdot)$, Vector Quantization Layer~$\quantizer(\cdot,\cdot)$  and Decoder~$\decoder(\cdot)$. Given a vector $\x$, the encoder maps it into a latent representation $\ze$. The codebook is a collection of $K$ code vectors, denoted as $\codebook=\{\c_1, \c_2, \ldots, \c_K\}$, where each vector $\c_i$ has same dimension as $\ze$. In quantization layer $\quantizer(\cdot,\cdot)$, $\ze$ is quantized into $\zq$ by assigning it into a code vector from codebook $\codebook$ using a distance measure $\d(\cdot,\cdot)$:
\begin{equation}
\label{eq:quant}
\centering 
\begin{split}
   \zq=\c_x, \quad \text{where} \quad x = \arg\min_{i}\d(\ze,\c_i).
\end{split}
\end{equation}
Therefore, VQNs can be denoted:
\begin{equation}
\begin{split}
      \haty 
     = \decoder(\quantizer(\encoder(\x),\codebook))
     =\decoder(\quantizer(\ze,\codebook))
     =\decoder(\zq).
\end{split}
\end{equation}
VQNs predict the output $\haty$ and compute loss with target $\y$ : $\loss(\y,\haty)$. The objective of VQNs is to optimize the following loss function:
\begin{equation}
\begin{split}
      \min_{\encoder,\quantizer,\decoder}\mathbb{E}_{(\x,\y)\sim \T}[\loss_{task}(\decoder(\quantizer(\encoder(\x),\codebook))), \y)],
\end{split}
\end{equation}
where $\T$ is training dataset. The extension to image reconstruction is straightforward by applying the formulation to 2D data; see Appendix~\ref{suppl:vq4IR} for details.

\textbf{Optimization Challenge.} Since the $\arg\min$ operation in the quantization layer $\quantizer(\cdot)$ is non-differentiable, a common approach is to apply the straight-through estimator (STE) during backpropagation~\citep{ste}, which allows gradients to bypass the discrete operation:
\begin{equation}
\begin{split}
      \zq = \ze + sg[(\zq - \ze)],
\end{split}
\end{equation}
where $sg[\cdot]$ is stop-gradient operation.
In addition, a commitment loss~\citep{vqvae} is introduced to ensure that the gradient flow on both sides of the quantization layer remains accurate:
\begin{equation}
\begin{split}
      \loss_{cmt}(\ze, \zq)=\d(\zq, sg[\ze])+\beta\d(\ze, sg[\zq]).
\end{split}
\end{equation}
Although this approach makes the objective differentiable and thus optimizable, it introduces several additional challenges. 


\textbf{\textit{Challenge 1:}} Straight-Through Estimation Bias.
The straight-through estimation error, defined as $\delta = \zq - \ze$, introduces bias into the training process. It directly affects the decoder $\decoder(\cdot)$ and indirectly impacts the encoder $\encoder(\cdot)$ through the commitment loss~\citep{straightening}.

\textbf{\textit{Challenge 2:}} One-step-behind Update. Through derivation (see Appendix~\ref{suppl:derivation_one}), we show that the commitment loss causes the model to align with historical representations instead of the current ones~\citep{robust,straightening}. Specifically, the codebook is updated as
\begin{equation}
\label{eq:old}
\begin{split}
    \zq^{(t+1)} \leftarrow (1 - \eta)\cdot \zq^{(t)} + \eta \cdot \ze^{(t)},
\end{split}
\end{equation}
and the decoder is trained on the previous codebook output.
This mismatch leads to a misalignment between the encoder and decoder.
Ideally, both should be updated using the current representation:
\begin{equation}
\label{eq:new}
\begin{split}
    \zq^{(t+1)} \leftarrow (1 - \eta)\cdot \zq^{(t)} + \eta \cdot \ze^{(t+1)}.
\end{split}
\end{equation}
\textbf{\textit{Challenge 3:}} Sparse Codebook Gradients.
The codebook $\codebook$ suffers from sparse gradients—only the selected code vectors receive updates, while the unselected ones get no gradient signal and are likely to remain unused indefinitely~\citep{cvq, ibq}.
This results in codebook collapse and limited representational capacity.

We use a toy experiment to illustrate the impact of the STE. As shown in Figure~\ref{fig:toy}, the optimization trajectories of $\ze$ and $\zq$ form a spiral shape, revealing hindered optimization (\textbf{Challenge 1} \& \textbf{2}). Moreover, the t-SNE visualization further shows that directly applying the STE leads to low usage, with most codebook entries left unused. This phenomenon arises from the sparse gradients received by the codebook (\textbf{Challenge 3}).


\section{Methodology}
\label{method}
\subsection{Preliminary Observations}

\begin{figure}[ht]
\begin{center}
    \includegraphics[width=1\linewidth]{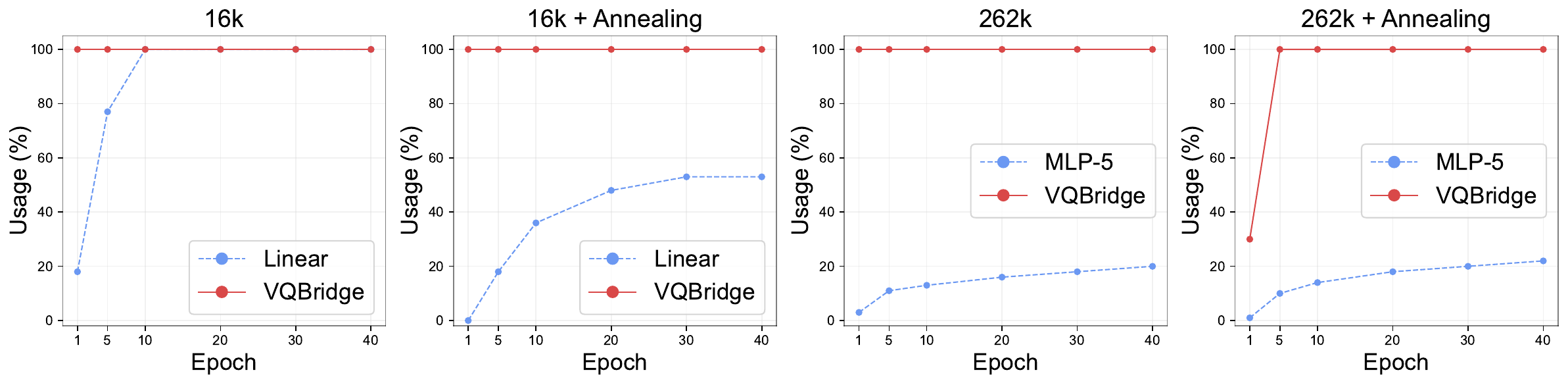}
\end{center}
\caption{\small
\textbf{Codebook usage during training.} With a codebook size of 16k, the linear-layer baseline can achieve 100\% usage, but usage drops significantly when learning annealing is applied. With a larger codebook size of 262k, even a 5-layer MLP fails to maintain sufficient usage. In contrast, VQBridge rapidly achieves 100\% usage under all configurations and with learning annealing, which greatly facilitates VQN optimization.
}
\label{fig:usage}
\end{figure}

We denote the corresponding sets of $\ze$, $\c$, and $\zq$ as $\zeset$, $\cset$, and $\zqset$, respectively. The distance between the distributions of the two sets can be formally defined as 
$$
D(\zeset,\cset)=\frac{1}{|\zeset|}\sum_{\mathbf{z}_i\in\zeset}\min_{\mathbf{c}_i\in\cset} d(\mathbf{z}_i, \mathbf{c}_i).
$$
Our work is based on the simple and effective linear reparameterization~\citep{straightening, vqganlc,simvq}, which can be generalized as a mapping function $f(\cdot)$ from $\cset$ to a new set $\hatcset$, with joint optimization over $\c$: $\hatcset = f(\cset)$. Consequently, the training objective shifts to minimizing the distance between $\zeset$ and $\hatcset$: $D(\zeset, \hatcset)$. A smaller $D(\zeset, \cset)$ indicates a smaller estimation error $\delta$. However, estimation error does not necessarily reflect optimization effectiveness, since a small number of active codebook vectors may lead to lower loss but fail to yield meaningful performance gains~\citep{straightening}. Codebook usage can therefore serve as a complementary metric, as higher usage indicates that more vectors are effectively contributing, which is more informative.
\begin{observation}\label{obs1}
To mitigate the straight-through estimation bias, rapid alignment of the two distributions is crucial, reducing $D(\zeset, \cset)$ early in training. Furthermore, estimation error must remain low for stable optimization. In practice, this requires using a larger codebook while maintaining high usage from early to late training. We present the effect of codebook size on the loss in Appendix~\ref{suppl:vqloss}.
\end{observation}

\begin{observation}\label{obs2}
Learning annealing is necessary, as a smaller learning rate helps reduce the one-step-behind effect and improves the overall alignment of the model. We establish an upper bound on the effect of the one-step-behind update and demonstrate that learning annealing serves as an effective remedy. Further derivations are presented in Appendix~\ref{suppl:effect_onestepbehind}.
\end{observation}

\begin{observation}\label{obs3}
Linear layer is weak. As shown in Figure~\ref{fig:usage}, we observe that when a linear layer is used as the mapping function  $f(\cdot)$, the codebook usage drops significantly during learning annealing. Additionally, under codebook scaling, the linear mapping struggles even more. Attempts to increase model capacity with a 5-layer MLP still result in poor usage. This suggests that the representational power of such mappings is inadequate for aligning the distributions of $\zeset$ and $\hatcset$.
\end{observation}

Based on the above observations, stable and scalable training of VQN requires a strong and robust mapping function $f(\cdot)$, combined with learning annealing. In the following section, we introduce our proposed design, referred to as \monitor{}.

\begin{figure}[ht]
\begin{center}
    \includegraphics[width=0.75\linewidth]{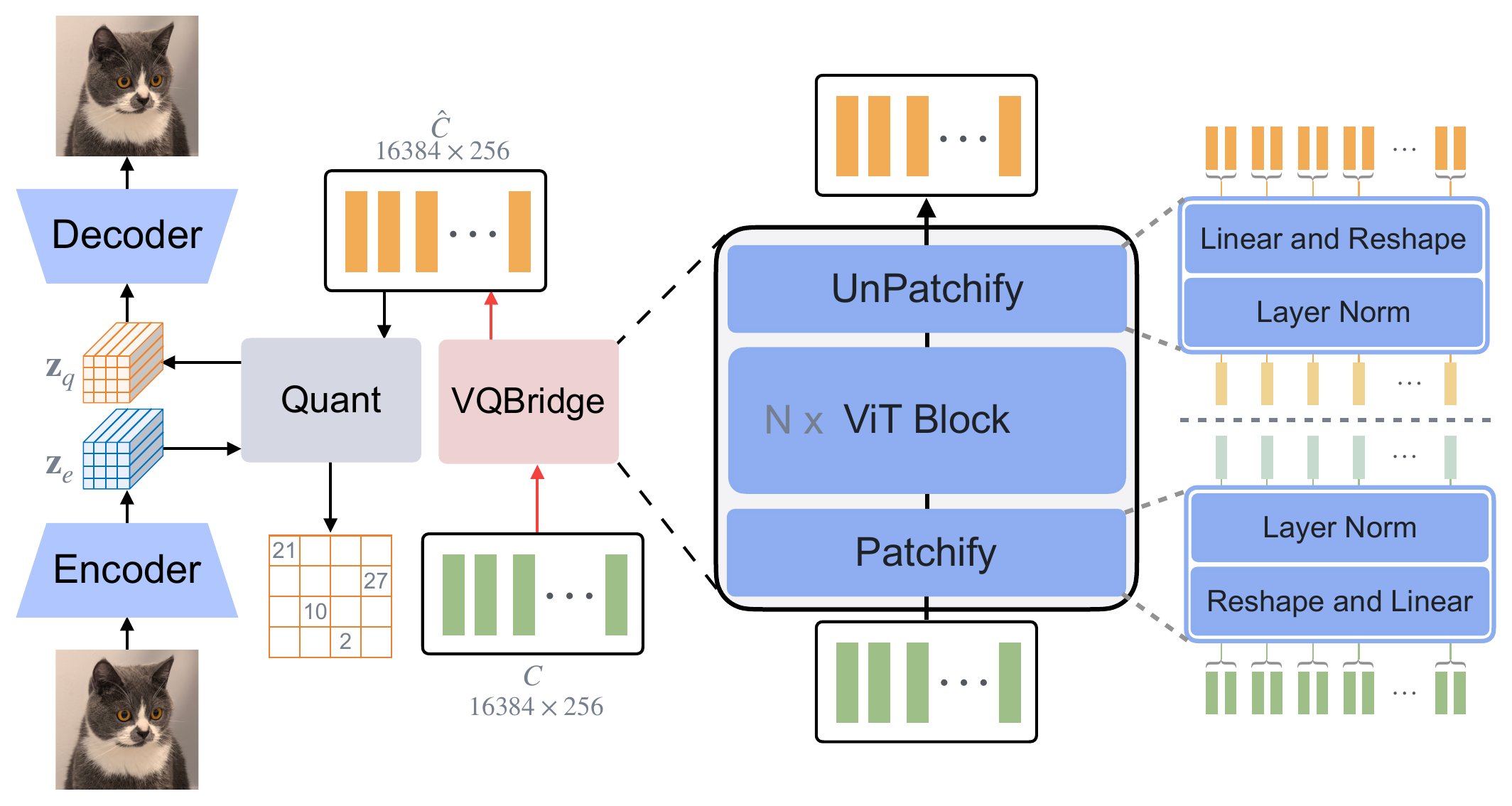}
\end{center}
\vspace{-4pt}
\caption{\small
\textbf{\model{} framwork overview.} \monitor{} first applies 1D patchification to the code vectors, then performs global interaction using ViT blocks, and finally restores them to their original size.
}
\label{fig:method}
\end{figure}
\vspace{-4pt}

\subsection{\monitor{} Design}
The design objective of \monitor{} is to achieve strong fitting capability, enabling rapid alignment between two distributions, while maintaining scalability and training stability. Inspired by the architecture of Diffusion Transformers (DiT)~\citep{dit}, the module follows a compress–process–recover pipeline, where the compression step enables efficient scaling to large codebooks (\eg, 262k entries). To simplify the problem, we assume that codebook $\codebook$ and the mapped codebook $\hatcodebook=f(\codebook)$ share the same size and vector channel. Details are provided in Figure~\ref{fig:method}.

First, we perform a 1D \emph{patchify} operation. Given a codebook $\codebook$ containing $K$ code vectors, we divide it into $p$ groups, each containing $\frac{K}{p}$ vectors.  
For each group, the vectors are compressed into a single vector via a shared linear projection $W_{\text{comp}} \in \mathbb{R}^{d \times d'}$, followed by a LayerNorm operation:
\begin{equation}
\mathbf{h}_g = \mathrm{LN}\big( \mathbf{C}_g W_{\text{comp}} \big), \quad g = 1, \dots, p.
\end{equation}

Next, $N$ ViT blocks are applied as scalable layers to process the intermediate representation $\mathbf{H} \in \mathbb{R}^{p \times d'}$, enabling global information exchange and allowing the model to fit complex relationships between code vectors:
\begin{equation}
\mathbf{H}' = \mathrm{ViT}^N(\mathbf{H}).
\end{equation}

Finally, a 1D \emph{unpatchify} operation restores the representation to its original size. Specifically, $\mathbf{H}'$ is first normalized via LayerNorm, then expanded to $p$ times its dimension through a shared linear projection $W_{\text{exp}} \in \mathbb{R}^{d' \times pd}$:
\begin{equation}
\mathbf{\tilde{H}} = \mathrm{LN}(\mathbf{H}') W_{\text{exp}},
\end{equation}
followed by a reshape operation that converts each expanded vector back into $p$ vectors within group:
\begin{equation}
\mathbf{\hat{C}} = \mathrm{reshape}(\mathbf{\tilde{H}}) \in \mathbb{R}^{K \times d}.
\end{equation}
This yields a codebook $\mathbf{\hat{C}}$ of the same size as the input to \monitor{}, where the code vectors have undergone full mutual interaction.

Note that both \monitor{} and the original codebook $\mathcal{C}$ can be discarded after training, maintaining only the mapped codebook $\hat{\mathcal{C}}$. This implies that \monitor{} introduces no additional computational overhead during inference, preserving the original VQN's workflow and computational cost. 





\section{Experiments}
\label{sec:experiments}

\subsection{Main Experiments}
\label{sec:main experiments}
\begin{table}[!ht]
\renewcommand\arraystretch{1.05}
\centering
\setlength{\tabcolsep}{2.3mm}{}
\small
{
\caption{
\textbf{Reconstruction performance of different tokenizers on $\boldsymbol{256 \times 256}$ ImageNet 50k validation set using the same Encoder and Decoder.} $^{\ast}$ trained on OpenImages~\citep{openimages}. $^{\dagger}$ trained for 330 epochs. $^{\ddagger}$ trained for 120 epochs. We highlight the best results in bold and underline the second-best results.
}\label{tab:reconstruction}
    \scalebox{1.0}
    {
        \begin{tabular}{l|ccc|ccc}
        \toprule
        \multirow{2}{*}{Method} & \multirow{2}{*}{Tokens} & Codebook & Vector & \multirow{2}{*}{rFID$\downarrow$} & \multirow{2}{*}{LPIPS$\downarrow$} & Codebook \\ 
         &  & Size & Channel& & & Usage$\uparrow$ \\
        \midrule
        VQGAN~\citep{vqgan}   & 16 $\times$ 16 &   1,024  & 256 & 7.94 & $-$ & 44\% \\
        VQGAN~\citep{vqgan}   & 16 $\times$ 16  &   16,384  & 256 & 4.98 & 0.17 & 5.9\% \\
        SD$^{\ast}$~\citep{ldm}   & 16 $\times$ 16  &   16,384 & 8 & 5.15 & $-$ & $-$ \\
        MaskGIT~\citep{maskgit}   & 16 $\times$ 16  &   1,024 & 256 & 2.28 & $-$ & $-$ \\
        LlamaGen~\citep{llamagen}   & 16 $\times$ 16  &   16,384 & 8 & 2.19 & 0.14 & 97$\%$ \\
        LlamaGen~\citep{llamagen}   & 16 $\times$ 16  &   16,384 & 256 & 9.21 & $-$ & 0.29$\%$ \\
        VQGAN-LC~\citep{vqganlc}   & 16 $\times$ 16  &   16,384 & 8 & 3.01 & \underline{0.13} & \underline{99$\%$} \\
        VQGAN-LC~\citep{vqganlc}   & 16 $\times$ 16  &   100,000 & 8 & 2.62 & \textbf{0.12} & \underline{99$\%$} \\
        IBQ$^{\dagger}$~\citep{ibq}   & 16 $\times$ 16  &   16,384 & 256 & 1.55 & 0.23 & 97$\%$ \\
        \midrule
        \model{}   & 16 $\times$ 16  & 16,384 & 256 & \underline{1.30} & \underline{0.13} & \textbf{100$\%$} \\
        \model{}$^{\ddagger}$   & 16 $\times$ 16  & 16,384 & 256 & \textbf{1.17} & \underline{0.13} & \textbf{100$\%$} \\
    \bottomrule
    \end{tabular}
    }
    \vspace{-2mm}
}
\end{table}
\begin{table}[!ht]
\renewcommand\arraystretch{1.05}
\vspace{-2mm}
\centering
\setlength{\tabcolsep}{2.5mm}{}
\small
{
\caption{
\textbf{Generative model comparison on class-conditional ImageNet $\boldsymbol{256 \times 256}$.} The metrics include Fréchet Inception Distance (FID), Inception Score (IS), Precision (Pre), and Recall (Rec). We highlight the best results in bold and underline the second-best results.
}\label{tab:generation}
    \scalebox{1.0}
    { 
        \begin{tabular}{c|l|c|cccc}
        \toprule
        Type & Model & \#Para & FID$\downarrow$ & IS$\uparrow$ & Pre$\uparrow$ & Rec$\uparrow$ \\
        \midrule
        Diff. & DiT-L/2~\citep{dit} & 458M & 5.02 & 167.2 &  0.75 & \underline{0.57} \\
        Diff. & DiT-XL/2~\citep{dit} & 675M & \underline{2.27} & 278.2 & \underline{0.83} & \underline{0.57} \\
        \midrule
        Mask. & MaskGIT~\citep{maskgit} & 227M & 6.18 & 182.1 & 0.80 & 0.51 \\
        \midrule
        VAR & VAR-$d16$~\citep{var} & 310M & 3.30 & 274.4 & \textbf{0.84} & 0.51 \\
        VAR & VAR-$d20$~\citep{var} & 600M & 2.57 & \textbf{302.6} & \underline{0.83} & 0.56 \\
        \midrule
        AR & VQGAN~\citep{vqgan} & 227M & 18.65 & 80.4 & 0.78 & 0.26 \\
        AR & LlamaGen-L~\citep{llamagen} & 343M & 3.81 & 248.3 & \underline{0.83} & 0.52 \\
        AR & LlamaGen-XL~\citep{llamagen} & 775M & 3.39 & 227.1 & 0.81 & 0.54 \\
        AR & IBQ-B~\citep{ibq} & 342M & 2.88 & 254.4 & \textbf{0.84} & 0.51 \\
        AR & IBQ-L~\citep{ibq} & 649M & 2.45 & 267.5 & 0.83 & 0.52 \\
        \midrule
        AR & \model{}-L (LlamaGen-L) & 343M & 2.39 & 276.6 & \textbf{0.84} & 0.56 \\
        AR & \model{}-XL (LlamaGen-XL) & 775M & \textbf{2.07} & \underline{287.0} & \underline{0.83} & \textbf{0.58} \\

    \bottomrule
    \end{tabular}
    }
    \vspace{-4mm}
}
\end{table}
\paragraph{Implementation Details.} 
For visual reconstruction, we adopt a codebook with a size of 16,384 and a vector channel of 256. \monitor{} is configured with a latent dimension of 256, a depth of 2 and a patch size of 16. The encoder and decoder configurations are identical to those in VQGAN~\citep{vqgan, llamagen}, utilizing 2 layers of ResBlocks. The model is trained for 40 epochs with a base learning rate of 1e-4 and a batch size of 128. Learning annealing starts with a 4-epoch warmup, followed by a learning rate decay to 1\% of the original value. The Adam optimizer~\citep{adam} is employed with $\beta_1 = 0.9$ and $\beta_2 = 0.95$. Finally, we scale the training time to 120 epochs. For visual generation, LlamaGen~\citep{llamagen} is utilized as the generator, with training configurations aligned with the open-source implementations. Additional details are provided in the Appendix~\ref{suppl:config}.


\textbf{Reconstruction Results.} In Table~\ref{tab:reconstruction}, we compare existing tokenizers using the same encoder and decoder architectures, most of which are trained on ImageNet-1k 256x256 with a downsampling ratio of 16. \model{} achieves an rFID of 1.30 after 40 epochs, outperforming all other discrete tokenizers. Scaling the training time to 120 epochs further improves the rFID to 1.17. By addressing the optimization challenges of vector quantization, our method can be extended to other discrete tokenizers, unlocking their full potential (see Section~\ref{sec:generalization}).

\begin{figure}[t]
\begin{center}
    \includegraphics[width=0.9\textwidth]{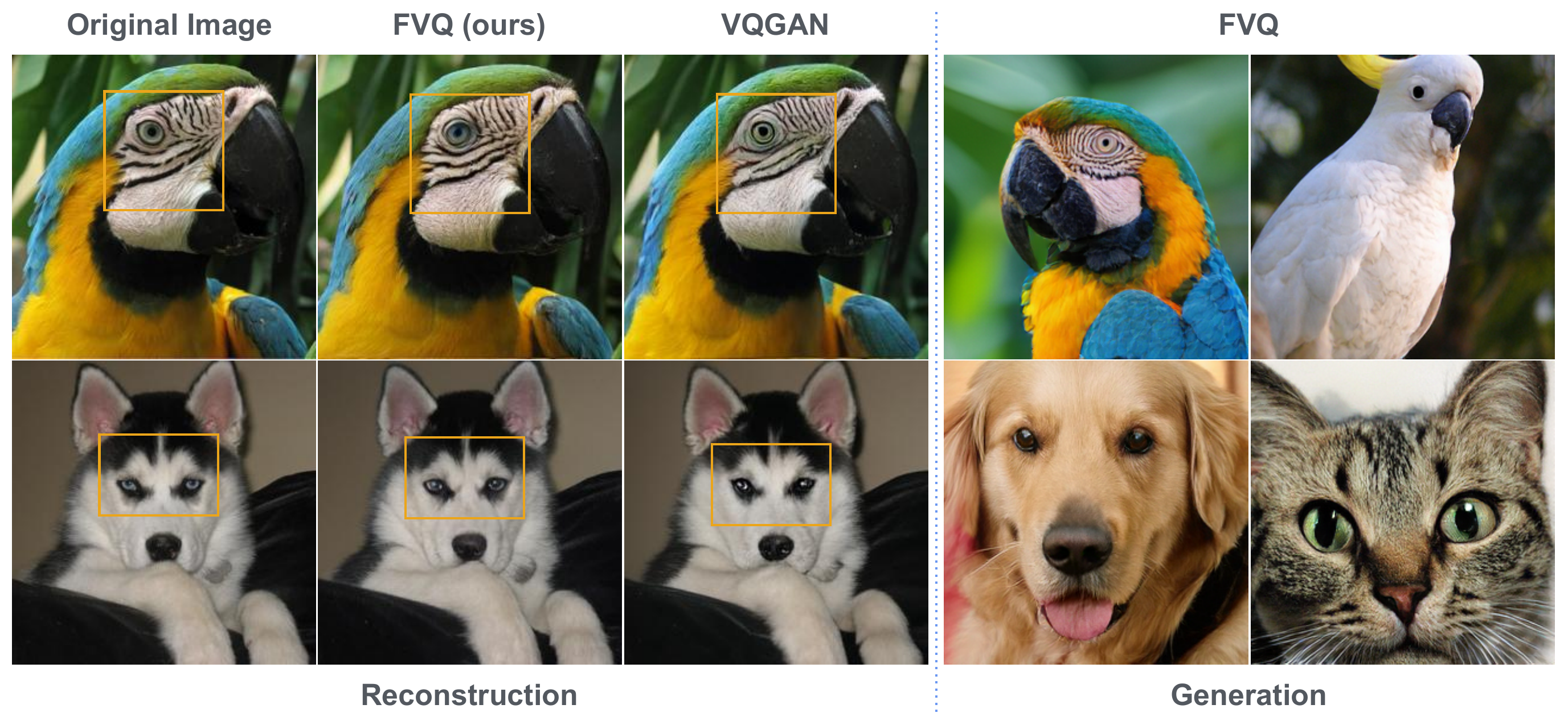}
\end{center}
\vspace{-4mm}
\caption{\small
\textbf{Qualitative results of \model.} \textbf{Left}: reconstruction performance compared against baseline methods. \textbf{Right}: generation results.
\vspace{-3mm}
}
\label{fig:vis}
\end{figure}

\textbf{Visual Generation.} We report the results of LlamaGen~\citep{llamagen} as generators in Table~\ref{tab:generation}, comparing them with prior generative models. \model{} improves the FID of LlamaGen-L from 3.81 to 2.39, surpassing other models with similar parameter sizes. Notably, \model{}-L~(343M) even outperforms VAR-$d20$~(600M, 2.39 \vs 2.57), and \model{}-XL exceeds DiT-XL/2~(2.07 \vs 2.27), demonstrating the potential of autoregressive (AR) methods. These results further emphasize that the tokenizer is the key to AR visual generation. Previous AR models are constrained by suboptimal tokenizers, whereas \model{} significantly enhances the generative capabilities of AR models, fully unlocking their potential. See Figure~\ref{fig:vis} for visualization results. 

\subsection{Ablation Studies}
\label{sec:preliminary experiments}
\begin{figure}[ht]
\centering
\includegraphics[width=1.0\textwidth]{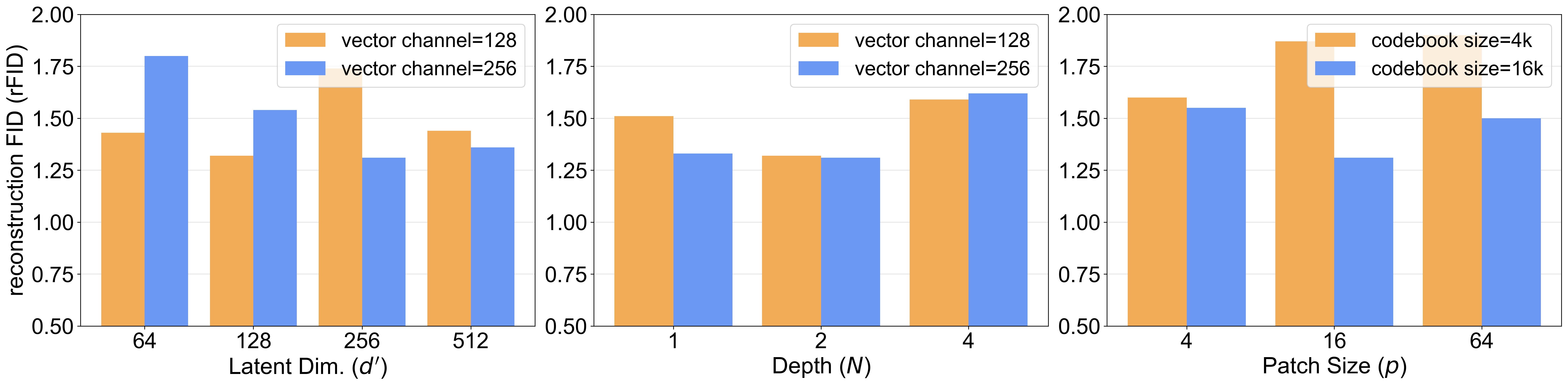}
\vspace{-3mm}
\caption{\textbf{\monitor{} desin ablation results.} We vary VQBridge configurations including the latent dimension ($d'$), ViT block depth ($N$), and patch size ($p$), as well as codebook parameters such as the vector channel and codebook size.}
\label{fig:pre}
\end{figure}

To identify the most efficient and effective setup, we explore various settings of \monitor{} in Figure~\ref{fig:pre}, focusing on the interplay between its parameters, codebook's size and vector channel.

\textbf{Latent Dimension ($d'$).} 
In left subfigure, increasing $d'$ from 64 to 256 consistently improves the reconstruction quality. Interestingly, the best results appear when the latent dimension $d'$ matches the vector channel size, indicating that a balanced latent embedding dimension and code vector width facilitates more effective representation learning. Further enlarging $d'$ to 512 does not provide additional benefits, and may even degrade performance due to over-parameterization and optimization difficulties. Therefore, we set $d'$ equal to the vector channel size in our final configuration.

\textbf{Depth of ViT Block ($N$).}
The middle subfigure shows that using a single ViT block leads to suboptimal results. Increasing the depth to $N=2$ significantly improves performance, while further stacking blocks ($N=4$) actually degrades performance due to increased optimization difficulty and higher computation cost. Therefore, $N=2$ is adopted as our default.

\textbf{Patch Size ($p$).} 
The right subfigure presents the effect of patch size under different codebook sizes, with both parameters following a $4^n$ scaling pattern. Our experiments reveal a clear scaling relationship: when $K=4$k, the optimal patch size is $p=4$, while when $K=16$k, the optimal patch size is $p=16$, indicating that patch size and codebook size should be scaled together to achieve optimal performance. This scaling relationship reflects a balance between the computational pressure on the Patchify stage and the ViT blocks.

Our final model adopts latent dimension $d'$ consistent with the vector channel dimension (default 256), 2 ViT blocks, and $p=16$ with a $K=16k$ codebook. When scaling up the codebook, we simultaneously increase the patch size by the same factor of $4\times$. This scaling strategy ensures optimal performance across different model configurations while preserving computational efficiency.

\subsection{Scaling Studies}
\label{sec:scaling studies}


Table~\ref{tab:scalability} presents a comprehensive evaluation of scalability across different model configurations, examining the effects of codebook size, vector channels, and training epochs on reconstruction quality and codebook usage. The results demonstrate performance differences between VQGAN and \model{} across multiple scaling dimensions. All models are trained on ImageNet~\citep{imagenet} and evaluated on both the ImageNet val set~(50k) and COCO 2017 val set~(5k)~\citep{laion_coco}.

\begin{table}[!ht]
\renewcommand\arraystretch{1.05}
\vspace{-2mm}
\centering
\setlength{\tabcolsep}{2.3mm}{}
\small
{
\caption{
\textbf{Results of scalability from codebook size, vector channel and training epoch.} rFID$_{\text{IN1K}}$ denotes the results evaluated on the ImageNet-1k validation set, while rFID$_{\text{COCO}}$ denotes the results evaluated on the COCO validation set. (-Ann.) denotes results without learning annealing.
}\label{tab:scalability}
    \scalebox{0.95}
    {
        \begin{tabular}{ll|cccc|ccc}
        \toprule
         & \multirow{2}{*}{Model} & {Training} & \multirow{2}{*}{Ratio} & {Codebook} & {Vector} & \multirow{2}{*}{rFID\textsubscript{IN1K}} & \multirow{2}{*}{rFID\textsubscript{COCO}} & \multirow{2}{*}{Usage} \\
    &   & Epoch &    & Size &  Channel  &       &     &       \\
        \midrule
        \graycell{1}  & VQGAN & 40  & 16 & 16k   & 8   & 2.19 & 8.43 & 97\%  \\
        \graycell{2}  & VQGAN & 40  & 16 & 16k   & 256 & 9.21 & $-$    & 0.29\% \\
        \midrule
        \graycell{3}  & \model{}$^{-{\rm Ann.}}$   & 40  & 16 & 16k   & 256 & 2.61 & $-$ & 100\% \\
        \midrule
        \ablanum{4}  & \model{}   & 40  & 16 & 16k   & 4 & 1.71 & 7.64 & 100\% \\
        \ablanum{5}  & \model{}   & 40  & 16 & 16k   & 8 & 1.47 & 7.15 & 100\% \\
        
        \ablanum{6}  & \model{}   & 40  & 16 & 16k   & 256 & 1.30 & 6.84 & 100\% \\
        \ablanum{7}  & \model{}   & 120 & 16 & 16k   & 256 & 1.17 & 6.62 & 100\% \\
        \ablanum{8}  & \model{}   & 40  & 16 & 65k   & 256 & 1.14 & 6.39 & 100\% \\
        \ablanum{9}  & \model{}   & 40  & 16 & 65k   & 512 & 1.09 & 6.38 & 100\% \\
        \ablanum{10}  & \model{}   & 120 & 16 & 65k   & 512 & 1.00 & 6.16 & 100\% \\
        \ablanum{11}  & \model{}   & 40  & 16 & 262k  & 256 & 0.95 & 6.04 & 100\% \\
        \ablanum{12}  & \model{}   & 120 & 16 & 262k  & 256 & 0.88 & 5.83 & 100\% \\
    \midrule
        \graycell{13} & VQGAN & 40  & 8  & 16k   & 8   & 0.98 & 5.03 & 97\%  \\
        \ablanum{14} & \model{}   & 40  & 8  & 16k   & 256 & 0.39 & 3.84 & 100\% \\
    \bottomrule
    \end{tabular}
    }
    \vspace{-2mm}
}
\end{table}
\textbf{Model Performance and Stability.} 
\model{} consistently outperforms VQGAN in both reconstruction quality and codebook usage. As shown in Figure~\ref{fig:usage}, \model{} achieves 100\% codebook usage early in training and maintains it throughout. Learning annealing further improves alignment in VQNs (from 2.61 to 1.30), as noted in \textbf{Observation~\ref{obs2}}, which is also confirmed by our experiments. \model{} sustains full codebook usage across all tested configurations, whereas VQGAN experiences severe codebook collapse in high-capacity settings, with usage dropping to only 0.29\% when employing 256 vector channels, validating \textbf{Observation~\ref{obs3}}. This stability advantage is crucial for practical applications, as it ensures reliable performance scaling without catastrophic failure modes.

\textbf{Codebook Size Scaling.} Our experiments reveal clear scaling benefits with increased codebook capacity. For \model{} with 40 training epochs and 256 vector channels, expanding the codebook from 16k to 262k entries yields substantial improvements in reconstruction quality, with rFID$_{\text{IN1K}}$ decreasing from 1.30 to 0.95 and rFID$_{\text{COCO}}$ from 6.84 to 6.04. This trend indicates that larger codebooks provide enhanced representational capacity without suffering from underutilization issues.

\textbf{Vector Channel Effects.} Increasing vector channels from 4 to 256 leads to substantial improvements in reconstruction quality across both datasets. On ImageNet, rFID$_{\text{IN1K}}$ decreases from 1.71 to 1.30, while on COCO, rFID$_{\text{COCO}}$ improves from 7.64 to 6.84. This trend demonstrates that higher-dimensional vector representations enable more expressive feature encoding, allowing the model to capture finer visual details and achieve better reconstruction fidelity. Importantly, these findings provide a crucial foundation for integrating a semantic tokenizer (\eg, CLIP~\citep{clip} with 512-dimensional embeddings) with vector quantization.

\textbf{Training Duration Impact.} 
Extending training from 40 to 120 epochs yields consistent but diminishing returns on ImageNet, while on COCO it results in steady improvements, with rFID$_{\text{COCO}}$ increasing by approximately 0.2 at each evaluation interval. 

\textbf{Compression Trade-offs.} 
Higher compression ratios (e.g., 16$\times$) make the reconstruction task more challenging, reducing reconstruction quality, while lower compression ratios (e.g., 8$\times$) facilitate learning and improve reconstruction fidelity. Both models benefit from reduced compression, but \model{} maintains its stability advantage across compression levels, achieving rFID$_{\text{IN1K}}$ of 0.39 and rFID$_{\text{COCO}}$ of 3.84 at 8$\times$ compression with full codebook usage.

These findings establish \model{} as a more scalable and reliable architecture for vector quantization tasks, demonstrating predictable performance improvements with increased computational resources while maintaining full codebook usage across diverse configurations.

\subsection{Generalization Studies}
\label{sec:generalization}

Recently, several multi-code representation methods have been proposed to improve VQ, where multiple tokens are used to represent a single spatial location (\eg, RQ-VAE~\citep{rqvae} and VAR~\citep{var}). To demonstrate the generalization ability of our approach, we further conduct experiments on these two methods.
\begin{table}[!ht]
\renewcommand\arraystretch{1.05}
\vspace{-3mm}
\centering
\setlength{\tabcolsep}{2.3mm}{}
\small
{
\caption{
\textbf{Multi-Code Representation VQ.} $680_{10step}$ denotes a code map constructed through 10 sequential quantization steps with step sizes 1, 2, 3, 4, 5, 6, 8, 10, 13, 16, where 680 is the sum of squares of the sequence. 
}\label{tab:generalization}
    \scalebox{1.0}
    {
        \begin{tabular}{l|ccc|ccc}
        \toprule
        \multirow{2}{*}{Method} & Training & Token & Codebook & Vector & \multirow{2}{*}{rFID$\downarrow$}  \\ 
         & Epoch & Map & Size & Channel&  \\
        \midrule
        RQ-VAE~\citep{rqvae} & 10 & $8\times 8\times 4$ & 16,384 & 256 & 4.73 \\
        RQ-VAE~\citep{rqvae} & 50 & $8\times 8\times 4$ & 16,384 & 256 & 3.20 \\
        \rowcolor{gray!20}
        \model{} (RQ-VAE) & 10 & $8\times 8\times 4$ & 16,384 & 256 & 2.98 \\
        \midrule
        VAR~\citep{var} & 40 & 680$_{10step}$ & 4,096 & 32 & 1.00 \\
        \rowcolor{gray!20}
        \model{} (VAR) & 40 & 680$_{10step}$ & 16,384 & 256 & 0.80 \\
    \bottomrule
    \end{tabular}
    }
    \vspace{-2mm}
}
\end{table}

We follow the original training protocols as closely as possible. The results are reported in Table~\ref{tab:generalization}, where our method consistently improves performance. Notably, with only 10 epochs of training, our approach surpasses RQ-VAE trained for 50 epochs. This highlights the strong generalization ability and training efficiency of our method.
\section{Conclusion}

We have systematically analyzed the challenges inherent in discrete tokenizers and identified the root causes of unstable training and codebook collapse. Based on these insights, we proposed \model{} with the \monitor{} projector, enabling robust, scalable, and high-quality VQ optimization. Our experiments show that high codebook usage and state-of-the-art reconstruction are achievable across diverse settings, and that strong tokenizers are essential for effective autoregressive image generation. Beyond improving current models, our work provides a foundation for extending VQ training to larger-scale and potentially unlabeled datasets. We believe these findings will inform future research on autoregressive visual generation and contribute to the development of more capable multimodal learning systems.

{
\small
\bibliography{iclr2026_conference}
\bibliographystyle{iclr2026_conference}
}

\clearpage
\appendix
\clearpage
\setcounter{page}{1}
\appendix

\section{Related Work}
\label{suppl:related work}
\paragraph{Image Quantization.} To convert image features into discrete tokens, VQ-VAE~\citep{vqvae} maps encoded features to the nearest vectors in a learned codebook. VQ-VAE2~\citep{vqvae2} improves quantization with a hierarchical encoding strategy, and VQGAN~\citep{vqgan} enhances perceptual quality via adversarial and perceptual losses. However, scaling the codebook size and vector channel often leads to codebook collapse—low usage and poor reconstruction. To mitigate this, ViT-VQGAN~\citep{vit-vqgan} reduces the vector channel and adds $\mathit{l}_2$ normalization to improve codebook usage. FSQ~\citep{fsq} and LFQ~\citep{magvit2} adopt similar low-channel designs to maintain usage with larger codebook. Reg-VQ~\citep{reg-vq} introduces prior regularization, while CVQ~\citep{cvq} resets underused codebook entries. Yet, these methods still struggle when both size and channel are scaled, limiting reconstruction quality. Recently, ~\citep{straightening} specifically analyzes the impact of the Straight-Through Estimator~\citep{ste}. IBQ~\citep{ibq} achieves high usage with large codebook and high channel via index backpropagation. VQGAN-LC~\citep{vqganlc} initializes codebook with pretrained features and employs a linear projector to jointly transform embedding. SimVQ~\citep{simvq} further highlights the importance of joint transformation and provides theoretical support for linear mappings. We argue that scaling both the codebook size and vector channel requires strong projector to fully optimize the codebook and ensure effective usage.
\paragraph{Image Generation.}
PixelCNN~\citep{pixelcnn} and iGPT~\citep{igpt} generates RGB pixels in a scanning order. VQGAN~\citep{vqgan} improves ~\citep{pixelcnn, igpt} by using an autoregressive approach in the latent space of VQVAE~\citep{vqvae}. VQVAE2~\citep{vqvae2} and RQ-Transformer~\citep{rqvae} adopt the same approach but with additional scales. MaskGIT~\citep{maskgit} employs masked prediction similar to BERT~\citep{bert}. LlamaGen~\citep{llamagen} demonstrates that autoregressive generation is possible even without visual bias, while VAR~\citep{var} explores visual autoregressive generation using multi-scale methods. RandAR~\citep{randar} trains transformer in random orders to improve performance. Additionally, diffusion models~\citep{ddpm, ddim, flowmatching, ldm, stablediffusion, dit} generate images by denoising from noise. MAR~\citep{mar} uses diffusion loss to eliminate the vector quantization process. 

\section{Limitations and Future Work}
\label{suppl:limitations}
We primarily focus on the VQVAE/VQGAN-based learning paradigm to validate the effectiveness of our method. The solutions to codebook collapse and unstable training are general and can be further applied to models like UniTok~\citep{unitok} for additional improvements. Additionally, we only use LlamaGen as the generator; employing other strategies, such as random order learning and parallel decoding, can further enhance the generative capabilities of AR models. 

\paragraph{Pretraining Tokenizer.} We plan to train more advanced discrete tokenizers on larger-scale datasets, such as OpenImages~\citep{openimages} and LAION-COCO~\citep{laion_coco}, to further enhance the reconstruction capabilities of discrete tokenizers and create better conditions for autoregressive generation.

\paragraph{Unified Tokenizer.} Our results demonstrate that the vector channel can be scaled to larger dimensions. This facilitates alignment with visual representation tokenizers like CLIP~\citep{clip} (dim=512), enabling a unified tokenizer to accomplish both generative and understanding tasks.

We will explore these aspects in the future.

\section{Additional Experiments in Observation}
\label{suppl:vqloss}
\begin{figure}[t]
\begin{floatrow}[2]
\figurebox{\caption{\textbf{Effect of codebook size on training loss}}}{%
  \includegraphics[width=0.45\textwidth]{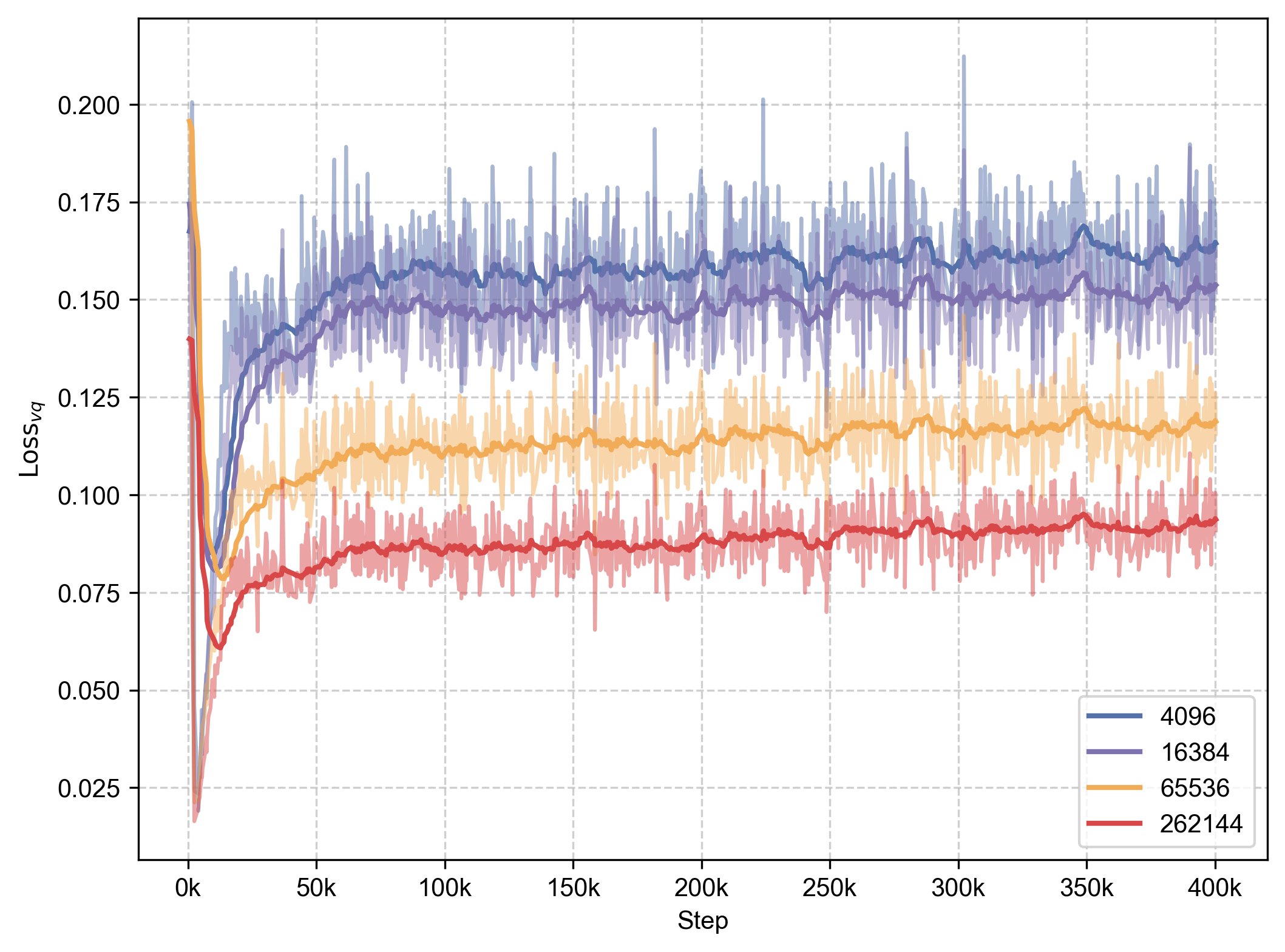}
  \label{fig:vqloss}
  }
\tablebox{\caption{\textbf{Additional observational results.} }}{%
  \label{tab:right}
  \scalebox{0.70}{%
        \begin{tabular}{l|cccc}
        \toprule
        Method & Codebook Size & usage & rFID  \\
        \midrule
        linear & 16k & 100\% & 2.10 \\
        linear+Annealing & 16k & 53\% & 1.52\\
        VQBridge & 16k & 100\% & 2.61  \\
        VQBridge+Annealing & 16k & 100\% & 1.30 \\
        \midrule
        MLP-5 & 262k & 20\% & 2.06 \\ 
        MLP-5+Annealing & 262k & 22\% & 1.75 \\
        VQBridge & 262k & 100\% & 1.40 \\
        VQBridge+Annealing & 262k & 100\% & 0.95 \\
        \bottomrule
        \end{tabular}
    }}
\end{floatrow}
\end{figure}

In Figure~\ref{fig:vqloss}, VQ loss tends to stabilize at a lower value as the codebook size increases. Early in training, the loss drops quickly because only a few code vectors are active. This allows the encoder to fit the codebook entries easily. Additional results are reported in Table~\ref{tab:right}, which demonstrate the necessity of using VQBridge together with annealing.

\floatsetup{heightadjust=all, floatrowsep=columnsep}
\newfloatcommand{figurebox}{figure}[\nocapbeside][\dimexpr(\textwidth-\columnsep)/2\relax]
\newfloatcommand{tablebox}{table}[\nocapbeside][\dimexpr(\textwidth-\columnsep)/2\relax]

\section{Derivation of the One-Step-Behind Formula}
\label{suppl:derivation_one}
Consider the $t$-th iteration of VQ training with the following components:
encoder output $\ze^{(t)} \in \mathbb{R}^d$,
quantized output $\zq^{(t)}$.

The Vector Quantization commitment loss is defined as:
\begin{equation}
\begin{split}
    \mathcal{L}_{\text{vq}}\left(\mathbf{z}_e^{(t)}, \mathbf{z}_q^{(t)}\right) = \frac{1}{2}\left\|\mathbf{z}_e^{(t)} - \mathbf{z}_q^{(t)}\right\|_2^2. 
\end{split}
\end{equation}
The gradient of the commitment loss with respect to the vector code $\mathbf{z}_q$ is:
\begin{equation}
\begin{split}
    \frac{\partial \mathcal{L}_{\text{vq}}\left(\mathbf{z}_e^{(t)}, \mathbf{z}_q^{(t)}\right)}{\partial \mathbf{z}_q^{(t)}} = \frac{1}{2} \cdot 2 \left(\mathbf{z}_q^{(t)} - \mathbf{z}_e^{(t)}\right).
\end{split}
\end{equation}
The update rule for $\mathbf{z}_q^{(t+1)}$ when optimizing with SGD is:
\begin{equation}
\begin{split}
    \mathbf{z}_q^{(t+1)} = \mathbf{z}_q^{(t)} - \eta \frac{\partial \mathcal{L}_{\text{vq}}\left(\mathbf{z}_e^{(t)}, \mathbf{z}_q^{(t)}\right)}{\partial \mathbf{z}_q^{(t)}}.
\end{split}
\end{equation}
Substituting the gradient:
\begin{equation}
\begin{split}
    \mathbf{z}_q^{(t+1)} = \mathbf{z}_q^{(t)} - \eta \cdot \left(\mathbf{z}_q^{(t)} - \mathbf{z}_e^{(t)}\right).
\end{split}
\end{equation}
Expanding and rearranging:
\begin{equation}
\label{eq:old1}
\begin{split}
    \mathbf{z}_q^{(t+1)} = \mathbf{z}_q^{(t)} - \eta \mathbf{z}_q^{(t)} + \eta \mathbf{z}_e^{(t)}
    = (1 - \eta) \cdot \mathbf{z}_q^{(t)} + \eta \cdot \mathbf{z}_e^{(t)}.
\end{split}
\end{equation}
From the above, we arrive at Equation~\ref{eq:old}.
The correct update is given by Equation~\ref{eq:new}, and we derive an executable formula following~\citep{straightening}. First,
\begin{equation}
\label{eq:t+1}
\begin{split}
    \mathbf{z}_e^{(t+1)} = \mathbf{z}_e^{(t)} - \eta\,\frac{\partial \mathcal{L}_{\text{task}}}{\partial \mathbf{z}_e}\Big|_t.
\end{split}
\end{equation}
Substituting Equation~\ref{eq:t+1} into Equation~\ref{eq:new}:
\begin{align}
\mathbf{z}_q^{(t+1)} &= (1 - \eta) \cdot \mathbf{z}_q^{(t)} + \eta \cdot \mathbf{z}_e^{(t+1)} \\
&= (1 - \eta) \cdot \mathbf{z}_q^{(t)} + \eta\left(\mathbf{z}_e^{(t)} - \eta\,\frac{\partial \mathcal{L}_{\text{task}}}{\partial \mathbf{z}_e}\Big|_t\right) \\
&= \underbrace{(1 - \eta) \cdot \mathbf{z}_q^{(t)} + \eta \cdot \mathbf{z}_e^{(t)}}_{\text{commitment loss}} - \eta^2\,\frac{\partial \mathcal{L}_{\text{task}}}{\partial \mathbf{z}_e}\Big|_t.
\end{align}
It can be observed that the additional term is multiplied by $\eta^2$. When annealing is applied, the contribution of this term is negligible, which underscores the importance of learning annealing.

\section{Effect of One-Step-Behind on Training Stability}
\label{suppl:effect_onestepbehind}
In the $(t+1)$-th forward pass, quantization uses the \emph{updated} codebook:
\begin{align}
x^{(t+1)} &= \arg\min_k \|\ze^{(t+1)} - \c_k^{(t)}\|_2^2, \\
\zq^{(t+1)} &= \c_{x^{(t+1)}}^{(t)}.
\end{align}
Comparing with the quantization point used for backpropagation in step $t$: $\zq^{(t)} = \c_{x^{(t)}}^{(t-1)}$, we observe a temporal misalignment: \textbf{Step $t$} encoder updates using signals around $\codebook^{(t-1)}$; \textbf{Step $t+1$ forward} quantization centers have changed to $\codebook^{(t)}$. This inconsistency constitutes \textbf{one-step-behind problem}.

\textbf{Mathematical Analysis.} The quantization center shift can be expressed as:
\begin{equation}
\underbrace{\zq^{(t+1)} - \zq^{(t)}}_{\text{Quantization center shift}} = \c_{x^{(t+1)}}^{(t)} - \c_{x^{(t)}}^{(t-1)}.
\label{eq:center_shift}
\end{equation}
This decomposes into two components:
\begin{equation}
= \underbrace{\big(\c_{x^{(t+1)}}^{(t)} - \c_{x^{(t+1)}}^{(t-1)}\big)}_{\text{Codebook update effect}} + \underbrace{\big(\c_{x^{(t+1)}}^{(t-1)} - \c_{x^{(t)}}^{(t-1)}\big)}_{\text{Cluster switching}}.
\end{equation}

By the triangle inequality,
\begin{equation}
\|\zq^{(t+1)}-\zq^{(t)}\|
\le \|\c_{x^{(t+1)}}^{(t)}-\c_{x^{(t+1)}}^{(t-1)}\|
+ \|\c_{x^{(t+1)}}^{(t-1)}-\c_{x^{(t)}}^{(t-1)}\|.
\end{equation}

\textbf{Bounded Change Analysis.}
Assume bounded changes:
\begin{equation}
\|\c_i^{(t)} - \c_i^{(t-1)}\| \leq \delta_c \quad \forall i, \quad
\|\ze^{(t+1)} - \ze^{(t)}\| \leq \delta_z.
\end{equation}

Define the assignment margin at time $t$ as the difference between the distances to the nearest and second-nearest centers (evaluated w.r.t.\ $\{\c_i^{(t-1)}\}$):
\begin{equation}
m^{(t)} := \|\ze^{(t)}-\c_{\text{2nd}}^{(t-1)}\| - \|\ze^{(t)}-\c_{\text{1st}}^{(t-1)}\|.
\end{equation}
Using the triangle inequality one obtains that if $m^{(t)}>2\delta_z$ then the nearest-center assignment is preserved (i.e.\ $x^{(t+1)}=x^{(t)}$); conversely, when $m^{(t)}\le 2\delta_z$ cluster switching may occur. In the no-switching case
\begin{equation}
\|\zq^{(t+1)}-\zq^{(t)}\|\le \delta_c,
\end{equation}
while in the worst case (arbitrary switching) we may only guarantee
\begin{equation}
\|\zq^{(t+1)}-\zq^{(t)}\|\le \delta_c + D_{\max},
\end{equation}
where $D_{\max}:=\max_{i,j}\|\c_i^{(t-1)}-\c_j^{(t-1)}\|$ is a (time-local) upper bound on inter-center distances.

Finally, if the task loss $\mathcal L$ is $L$-Lipschitz w.r.t.\ $\zq$, then
\begin{equation}
\big|\mathcal L(\zq^{(t+1)}) - \mathcal L(\zq^{(t)})\big|
\le L\|\zq^{(t+1)}-\zq^{(t)}\|.
\end{equation}
Combining the above yields the following interpretation: when encoder outputs lie away from cluster boundaries ($m^{(t)}>2\delta_z$) and codebook updates are small ($\delta_c$ small), loss fluctuations induced by one-step-behind are controlled by $L\delta_c$; however, near boundaries cluster switching can produce center jumps of order inter-cluster distances, which the Lipschitz property amplifies into large loss fluctuations and thus destabilizes training.

\textbf{Implication for the Decoder.}
Since the quantized vector $\zq$ is directly fed into the decoder, fluctuations in $\zq$ propagate to the reconstruction stage. In the stable case (no switching, small $\delta_c$), the decoder receives smoothly varying inputs and can learn a consistent mapping. In contrast, when cluster switching occurs, $\zq$ may undergo discontinuous jumps of order $D_{\max}$, forcing the decoder to fit inconsistent latent representations across consecutive steps. This misalignment not only destabilizes optimization but also degrades reconstruction quality, as the decoder struggles to reconcile abrupt changes in its input space.





\section{Vector Quantization for Image Reconstruction}
\label{suppl:vq4IR}
A discrete tokenizer contains Encoder, Codebook and Decoder. Given an image $\mathbf{I} \in \mathbb{R}^{H\times W \times 3}$, the Encoder compresses it into an image feature map $\mathbf{Z}\in \mathbb{R}^{h\times w \times D}$, where $H\times W$ and $h\times w$ denote the spatial resolution of the image and feature map, $D$ denotes the channel number of feature map. Formally,
\begin{equation}
\label{eq:encoder}
\begin{split}
    \mathbf{Z} = \mathrm{Encoder}(\mathbf{I}).
\end{split}
\end{equation}
 The Codebook contains $K$ vectors $\mathcal{C}=\{\mathbf{c}_1, \mathbf{c}_2, ..., \mathbf{c}_K\}$, where each vector $\mathbf{c}_i$ is a $D$-dimensitional vector. By searching for the most similar vector in the codebook, $\mathbf{Z}_e$ is mapped to a feature $\mathbf{Z}_q$, and the indices form the token map $\mathbf{T}$. We denote the search process as $\mathrm{Quant}(\cdot, \cdot)$. We have
\begin{equation}
\label{eq:quantizer}
\centering 
\begin{split}
   \mathbf{c}_x = \mathrm{Quant}(\mathbf{c}^{i,j}, \mathcal{C}),
\end{split}
\end{equation}
where $ x=\arg\min_{k\in \{1, 2,...,K\}} \|\mathbf{z}_e^{i, j} - \mathbf{c}_k\|_2$, and 
\begin{equation}
\label{eq:Z_hat}
\centering 
\begin{split}
    \mathbf{Z}_q = \{\mathbf{c}_t^{i, j}\}, 
    \mathbf{X} = \{x^{i, j}\},
\end{split}
\end{equation}
where $i \in \{1, ..., h\}, j \in \{1, ..., w\}$.
The Decoder reconstructs $\mathbf{Z}_q$ into $\mathbf{\hat{I}}$. Formally, 
\begin{equation}
\label{eq:decoder}
\begin{split}
    \mathbf{\hat{I}}=\mathrm{Decoder}(\mathbf{\hat{Z}}), 
\end{split}
\end{equation}
Typically, a compound loss $\mathcal{L}$ is optimized to train the model. We have 
\begin{equation}
\label{eq:loss}
\begin{split}
    \mathcal{L}=\mathcal{L}_{rec} + \mathcal{L}_{commit} + \lambda_{perc}\mathcal{L}_{perc} + \lambda_{G}\mathcal{L}_G,
\end{split}
\end{equation}
where $\mathcal{L}_{rec}=\|\mathbf{I}-\mathbf{\hat{I}}\|_2$, 
$\mathcal{L}_{commit}=\|sg[\mathbf{Z}_e]-\mathbf{Z}_q\|_2 + \beta\|sg[\mathbf{Z}_q]-\mathbf{Z}_e\|_2$, $\mathcal{L}_{perc}$ is a perceptual loss like LPIPS~\citep{lpips}, $\mathcal{L}_G$ is a discriminative loss like PatchGAN~\citep{patchgan}, $\beta$, $\lambda_{perc}$, $\lambda_{G}$ are loss weights, and $sg[\cdot]$ is stop-gradient operation. 

\section{Further Details on Experimental Configurations}
\label{suppl:config}
\begin{table}[ht]
\renewcommand\arraystretch{1.2}
\centering
\setlength{\tabcolsep}{4mm}

\small
{
\caption{
\textbf{Experimental Configurations} on reconstruction and generation. Warmup(0.1) indicates that 10\% of the training time is used for learning rate warmup.
}\label{tab:configuration}
\scalebox{1.0}{
    \begin{tabular}{l|c|c}
    \toprule
    Config & Reconstruction & Generation \\ 
    \midrule
        Base Batch Size & 128 & 256 \\ 
        Training Epochs & 40 or 120 & 300--350 \\
        Optimizer & Adam & AdamW \\
        Base Learning Rate & $1 \times 10^{-4}$ & $1 \times 10^{-4}$ \\
        Learning Rate Schedule & warmup(0.1) & warmup(0) \\
        & [baselr(0.3)--linear(0.7)] & [baselr(0.05)--linear(0.95)] \\
        Warmup Learning Rate & 0.005 to 1 & $-$ \\
        Linear Learning Rate & 1 to 0.01 & 1 to 0.01 \\
        Optimizer Parameters & $\beta_1=0.9, \beta_2=0.95$ & $\beta_1=0.9, \beta_2=0.95$ \\
        & & $weight\_decay=5 \times 10^{-2}$ \\
        Loss Weight & commit loss $\beta=0.25$ & \\
        & perceptual loss $\lambda_{perc}=1.0$ & \\
        & discriminative loss $\lambda_G=0.5$ &  \\
    \bottomrule
    \end{tabular}
    }
}
\end{table}
The training recipe Table~\ref{tab:configuration} follows LlamaGen~\citep{llamagen}, where the learning rate is calculated as (batch size / base batch size) $\times$ base learning rate. The learning rate schedule specifies the proportions of different phases: warmup (0.1) indicates 10\% of the training epochs, and [baselr(0.3)-linear(0.7)] represents the remaining 90\%, with 30\% maintaining a constant learning rate and 70\% applying linear decay.
All training is conducted on A100 GPUs, and evaluation follows the LlamaGen protocol. We use Fréchet Inception Distance (FID) as the primary metric, sampling 50,000 images and evaluating FID using code from ADM~\citep{adm}. Additionally, we report Inception Score (IS), Precision, and Recall.


\section{Further Results Comparison}
\label{suppl:further_results}
\paragraph{Additional reconstruction results.}
We present results for more metrics, including RSNR and SSIM, in Table~\ref{tab:more_metric}. We report additional comparison results in Table~\ref{tab:all_reconstruction}, highlighting the superiority of \model{}. As shown in the Table~\ref{tab:all_reconstruction}, \model{} achieves the best reconstruction performance (rFID=0.88) among all discrete tokenizers, outperforming Open-MAGVIT2 and IBQ, even though they utilize larger architectures. 
\begin{table}[!th]
\renewcommand\arraystretch{1.05}
\centering
\setlength{\tabcolsep}{2.3mm}{}
\small
{
\caption{
\textbf{Additional Metrics and Comparisons.} $^{\ddagger}$ trained for 120 epochs. 
}\label{tab:more_metric}
    \scalebox{1.0}
    {
        \begin{tabular}{l|ccccc}
        \toprule
        Method & Codebook Size & rFID$\downarrow$ & RSNR$\uparrow$ & SSIM$\uparrow$ & Usage \\
        \midrule
        LlamaGen~\citep{llamagen} & 16,384 & 2.19 & 20.79 & 0.675 & 97.0\% \\
        \model{} & 16,384 & 1.30 & 22.36 & 0.648 & 100$\%$ \\
        \model{}$^{\ddagger}$ & 16,384 & 1.17 & 22.45 & 0.653 & 100$\%$ \\
        \model{} & 262,144 & 0.95 & 22.90 & 0.665 & 100$\%$ \\
        \model{}$^{\ddagger}$ & 262,144 & 0.88 & 22.95 & 0.672 & 100$\%$ \\
    \bottomrule
    \end{tabular}
    }
}
\end{table}
\begin{table}[!th]
\renewcommand\arraystretch{1.05}
\centering
\setlength{\tabcolsep}{2.3mm}{}
\small
{
\caption{
\textbf{Reconstruction performance of different tokenizers on $\boldsymbol{256 \times 256}$ ImageNet 50k validation set.} $^{\ast}$ trained on OpenImages~\citep{openimages}. $^{\dagger}$ trained for 330 epochs. $^{\ddagger}$ trained for 120 epochs. $^{\diamond}$ trained with 4 ResBlocks. 
}\label{tab:all_reconstruction}
    \scalebox{0.90}
    {
        \begin{tabular}{l|ccc|ccc}
        \toprule
        \multirow{2}{*}{Method} & \multirow{2}{*}{Tokens} & Codebook & Vector & \multirow{2}{*}{rFID$\downarrow$} & \multirow{2}{*}{LPIPS$\downarrow$} & Codebook \\ 
         &  & Size & Channel& & & Usage$\uparrow$ \\
        \midrule
        VAE-SD$^{\ast}$~\citep{ldm} & 16 $\times$ 16 &  & 16 & 0.87 & $-$ & $-$ \\
        VAE-MAR~\citep{mar}   & 16 $\times$ 16  &   $-$ & 16 & 1.22 & $-$ & $-$ \\
        \midrule
        \rowcolor{gray!20}
        \multicolumn{7}{c}{\textbf{\textit{VQ-VAE-based Methods}}}\\
        VQGAN~\citep{vqgan}   & 16 $\times$ 16  &   1,024  & 256 & 7.94 & $-$ & 44\% \\
        VQGAN~\citep{vqgan}   & 16 $\times$ 16  &   16,384  & 256 & 4.98 & 0.17 & 5.9\% \\
        VQGAN-SD$^{\ast}$~\citep{ldm}   & 16 $\times$ 16  &   16,384 & 8 & 5.15 & $-$ & $-$ \\
        MaskGIT~\citep{maskgit}   & 16 $\times$ 16  &   1,024 & 256 & 2.28 & $-$ & $-$ \\
        LlamaGen~\citep{llamagen}   & 16 $\times$ 16  &   16,384 & 8 & 2.19 & 0.14 & 97$\%$ \\
        LlamaGen~\citep{llamagen}   & 16 $\times$ 16  &   16,384 & 256 & 9.21 & $-$ & 0.29$\%$ \\
        VQGAN-LC~\citep{vqganlc}   & 16 $\times$ 16  &   16,384 & 8 & 3.01 & 0.13 & 99$\%$ \\
        VQGAN-LC~\citep{vqganlc}   & 16 $\times$ 16  &   100,000 & 8 & 2.62 & 0.12 & 99$\%$ \\
        Open-MAGVIT2$^{\dagger\diamond}$~\citep{open-magvit2} & 16 $\times$ 16  &   16,384 & 0 & 1.58 & 0.23 & 100$\%$ \\
        Open-MAGVIT2$^{\dagger\diamond}$~\citep{open-magvit2} & 16 $\times$ 16  &  262,144 & 0 & 1.17 & 0.20 & 100$\%$ \\
        IBQ$^{\dagger}$~\citep{ibq}  & 16$\times$ 16 & 16,384 & 256 & 1.55 & 0.23 & 97$\%$ \\
        IBQ$^{\dagger\diamond}$~\citep{ibq} & 16 $\times$ 16  & 16,384 & 256 & 1.37 & 0.22 & 96$\%$ \\
        IBQ$^{\dagger\diamond}$~\citep{ibq} & 16 $\times$ 16  & 262,144 & 256 & 1.00 & 0.20 & 84$\%$ \\
        \midrule
        \model{} & 16 $\times$ 16  & 16,384 & 256 & 1.30 & 0.13 & 100$\%$ \\
        \model{}$^{\ddagger}$ & 16 $\times$ 16  & 16,384 & 256 & 1.17 & 0.13 & 100$\%$ \\
        \model{} & 16 $\times$ 16  & 262,144 & 256 & 0.95 & 0.12 & 100$\%$ \\       
        \model{}$^{\ddagger}$ & 16 $\times$ 16  & 262,144 & 256 & 0.88 & 0.12 & 100$\%$ \\ 
        \midrule
        \rowcolor{gray!20}
        \multicolumn{7}{c}{\textbf{\textit{1-D Token Sequence}}}\\
        Titok-L~\citep{titok} & 32 & 4,096 & 16 & 2.21 &$-$&$-$ \\
        Titok-B~\citep{titok} & 64 & 4,096 & 16 & 1.70 & $-$&$-$ \\
        Titok-S~\citep{titok} & 128 & 4,096 & 16 & 1.71 & $-$&$-$\\
        \midrule
        \rowcolor{gray!20}
        \multicolumn{7}{c}{\textbf{\textit{Multi-Code Representation}}}\\
        RQ-VAE$_{10ep}$~\citep{rqvae} & 8 $\times$ 8 $\times$ 4&16,384&256&4.73&$-$&$-$ \\
        RQ-VAE$_{50ep}$~\citep{rqvae} & 8 $\times$ 8 $\times$ 4&16,384&256&3.20&$-$&$-$ \\
        MoVQ~\citep{movq} & 16 $\times$ 16 $\times$ 4 & 1,024 & 64 & 1.12 & $-$ & $-$ \\
        VAR~\citep{var} & 680$_{10step}$ & 4,096 & 32 & 1.00 & $-$ & $-$ \\
        \midrule
        \model{} (RQ-VAE$_{10ep}$) & 8 $\times$ 8 $\times$ 4 & 16,384 & 256 & 2.98 & $-$ & $-$ \\
        \model{} (VAR) & 680$_{10step}$ & 16,384 & 256 & 0.80 & $-$ & $-$ \\
    \bottomrule
    \end{tabular}
    }
}
\end{table}

\clearpage
\paragraph{Additional ablation experiments.}
We perform addtional ablation experiments to study the effects of warmup and learning annealing. 
Our results indicate that warmup considerably influences performance, in line with the observations reported in~\citep{straightening}. 
In contrast, the choice of learning annealing schedule—linear or cosine—has a negligible impact on the outcome. 
The detailed results are summarized in Table~\ref{tab:more_ablation}. 	Table~\ref{tab:memory_usage} reports the computational cost of different projectors, highlighting the efficiency of VQBridge.

\begin{table}[!th]
\centering
\renewcommand\arraystretch{1.05}
\setlength{\tabcolsep}{2.3mm}
\vspace{-1mm}
\small
\begin{floatrow}
\tablebox{%
\scalebox{0.85}{
\begin{tabular}{l|cc}
\toprule
Method & Codebook Size & rFID \\
\midrule
baseline w/ linear & 16,384 & 1.30 \\
w/o warmup & 16,384 & 1.70 \\
w/ cosine & 16,384 & 1.32 \\
\bottomrule
\end{tabular}
}
}{%
\caption{\textbf{Additional Ablation Experiments.}}
\label{tab:more_ablation}
}
\tablebox{%
\scalebox{0.85}{
\begin{tabular}{cccc}
\toprule
Codebook Size & Linear & MLP-5 & VQBridge \\
\midrule
16,384   & 1.07G  & 5.37G  & 4.30G  \\
262,144  & 17.18G & 85.90G & 55.89G \\
\bottomrule
\end{tabular}
}
}{%
\caption{\textbf{Estimated Model Computational Cost (in FLOPs)}}
\label{tab:memory_usage}
}
\end{floatrow}
\vspace{-2mm}
\end{table}




\paragraph{Additional generative results.}
We report more additional generative results in Table~\ref{tab:all_generation}. \model{}-XL~(775M, 2.07) surpasses VAR-d24~(1.0B, 2.09), RandAR-XXL~(1.4B, 2.15) and L-DiT~(3.0B, 2.10; 7.0B, 2.28), further demonstrating the importance of tokenizers and the superiority of AR methods. \model{} can also further enhance the capabilities of other AR models. Due to time and computational constraints, we plan to extend this work in the future.
\begin{table}[!th]
\renewcommand\arraystretch{1.05}
\centering
\setlength{\tabcolsep}{2.5mm}{}
\small
{
\caption{
\textbf{Generative model comparison on class-conditional ImageNet $\boldsymbol{256 \times 256}$.} The metrics include Fréchet Inception Distance (FID), Inception Score (IS), Precision (Pre), and Recall (Rec). $^{\ddagger}$ trained with a learning rate schedule.
}\label{tab:all_generation}
    \scalebox{0.95}
    { 
        \begin{tabular}{c|l|c|cccc}
        \toprule
        Type & Model & \#Para & FID$\downarrow$ & IS$\uparrow$ & Pre$\uparrow$ & Rec$\uparrow$ \\
        \midrule
        GAN & BigGAN~\citep{biggan} & 112M & 6.95 & 224.5 & 0.89 & 0.38 \\
        GAN & GigaGGAN~\citep{gigagan} & 569M &3.45 & 225.5 & 0.84 & 0.61 \\
        GAN & StyleGan-XL~\citep{stylegan-xl} & 166M & 2.30 & 265.1 & 0.78 & 0.53 \\
        \midrule
        Diff. & ADM~\citep{adm} & 554M & 10.94 & 101.0 & 0.69 & 0.63 \\
        Diff. & CDM~\citep{cdm} & $-$ & 4.88 & 158.7 & $-$ & $-$ \\
        Diff. & LDM-4-G~\citep{ldm} & 400M & 3.60 & 247.7 & $-$ & $-$ \\
        Diff. & DiT-L/2~\citep{dit} & 458M & 5.02 & 167.2 &  0.75 & 0.57 \\
        Diff. & DiT-XL/2~\citep{dit} & 675M & 2.27 & 278.2 & 0.83 & 0.57 \\
        Diff. & L-DiT-3B~\citep{largedit} & 3.0B & 2.10 & 304.4 & 0.82 & 0.60 \\
        Diff. & L-DiT-7B~\citep{largedit} & 7.0B & 2.28 & 316.2 & 0.83 & 0.58 \\
        \midrule
        Mask. & MaskGIT~\citep{maskgit} & 227M & 6.18 & 182.1 & 0.80 & 0.51 \\
        Mask. & RCG~\citep{rcg} & 502M & 3.49 & 215.5 & $-$ & $-$ \\
        \midrule
        MAR & MAR-L~\citep{mar} & 479M & 1.98 & 290.3 & $-$ & $-$ \\
        MAR & MAR-H~\citep{mar} & 943M & 1.55 & 303.7 & 0.81 & 0.62 \\
        \midrule
        VAR & VAR-$d16$~\citep{var} & 310M & 3.30 & 274.4 & 0.84 & 0.51 \\
        VAR & VAR-$d20$~\citep{var} & 600M & 2.57 & 302.6 & 0.83 & 0.56 \\
        VAR & VAR-$d24$~\citep{var} & 1.0B & 2.09 & 321.9 & 0.82 & 0.59 \\
        VAR & VAR-$d30$~\citep{var} & 2.0B & 1.92 & 323.1 & 0.82 & 0.59 \\
        \midrule
        AR & VQGAN~\citep{vqgan} & 227M & 18.65 & 80.4 & 0.78 & 0.26 \\
        AR & ViTVQ~\citep{vit-vqgan} & 1.7B & 4.17 & 175.1 & $-$ & $-$ \\
        AR & RQTran.~\citep{rqvae} & 3.8B & 7.55 & 134.0 & $-$ & $-$ \\
        AR & RandAR-L~\citep{randar} & 343M & 2.55 & 288.9 & 0.81 & 0.58 \\
        AR & RandAR-XL~\citep{randar} & 775M & 2.25 & 317.8 & 0.80 & 0.60 \\
        AR & RandAR-XXL~\citep{randar} & 1.4B & 2.15 & 321.9 & 0.79 & 0.60 \\
        AR & LlamaGen-L~\citep{llamagen} & 343M & 3.81 & 248.3 & 0.83 & 0.52 \\
        AR & LlamaGen-XL~\citep{llamagen} & 775M & 3.39 & 227.1 & 0.81 & 0.54 \\
        AR & LlamaGen-XXL~\citep{llamagen} & 1.4B & 3.09 & 253.6 & 0.83 & 0.53 \\
        AR & IBQ-B~\citep{ibq} & 342M & 2.88 & 254.4 & 0.84 & 0.51 \\
        AR & IBQ-L~\citep{ibq} & 649M & 2.45 & 267.5 & 0.83 & 0.52 \\
        AR & IBQ-XL~\citep{ibq} & 1.1B & 2.14 & 279.0 & 0.83 & 0.56 \\
        \midrule
        AR & \model{}-L (LlamaGen-L) & 343M & 2.39 & 275.8 & 0.84 & 0.56 \\
        AR & \model{}-XL (LlamaGen-XL) & 775M & 2.07 & 287.0 & 0.83 & 0.58 \\
    \bottomrule
    \end{tabular}
    }
}
\end{table}


\definecolor{orange_distribution}{HTML}{f79674}
\definecolor{reddot}{HTML}{D94848}
\definecolor{greendot}{HTML}{8AA66F}

\clearpage

\section{Further Visualization Analysis}
\label{suppl:visualization}
\begin{figure}[htb]
\begin{center}
    \includegraphics[width=1.0\textwidth]{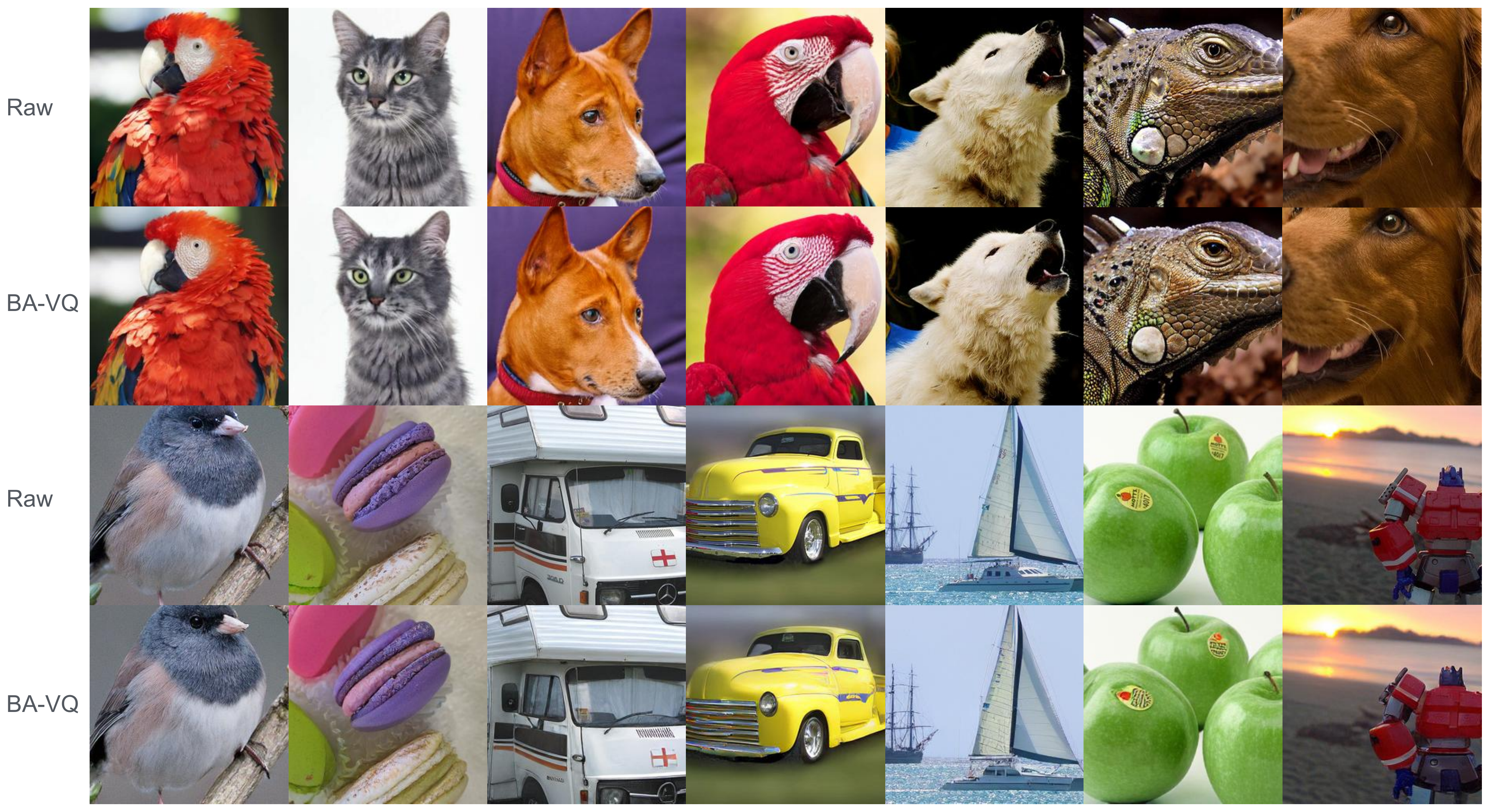}
\end{center}
\caption{\small
\textbf{\model{}-16,384~(rFID=1.17) reconstruction samples.} 
}
\label{fig:recon-demo}
\end{figure}

\begin{figure}[htb]
\begin{center}
    \includegraphics[width=1.0\textwidth]{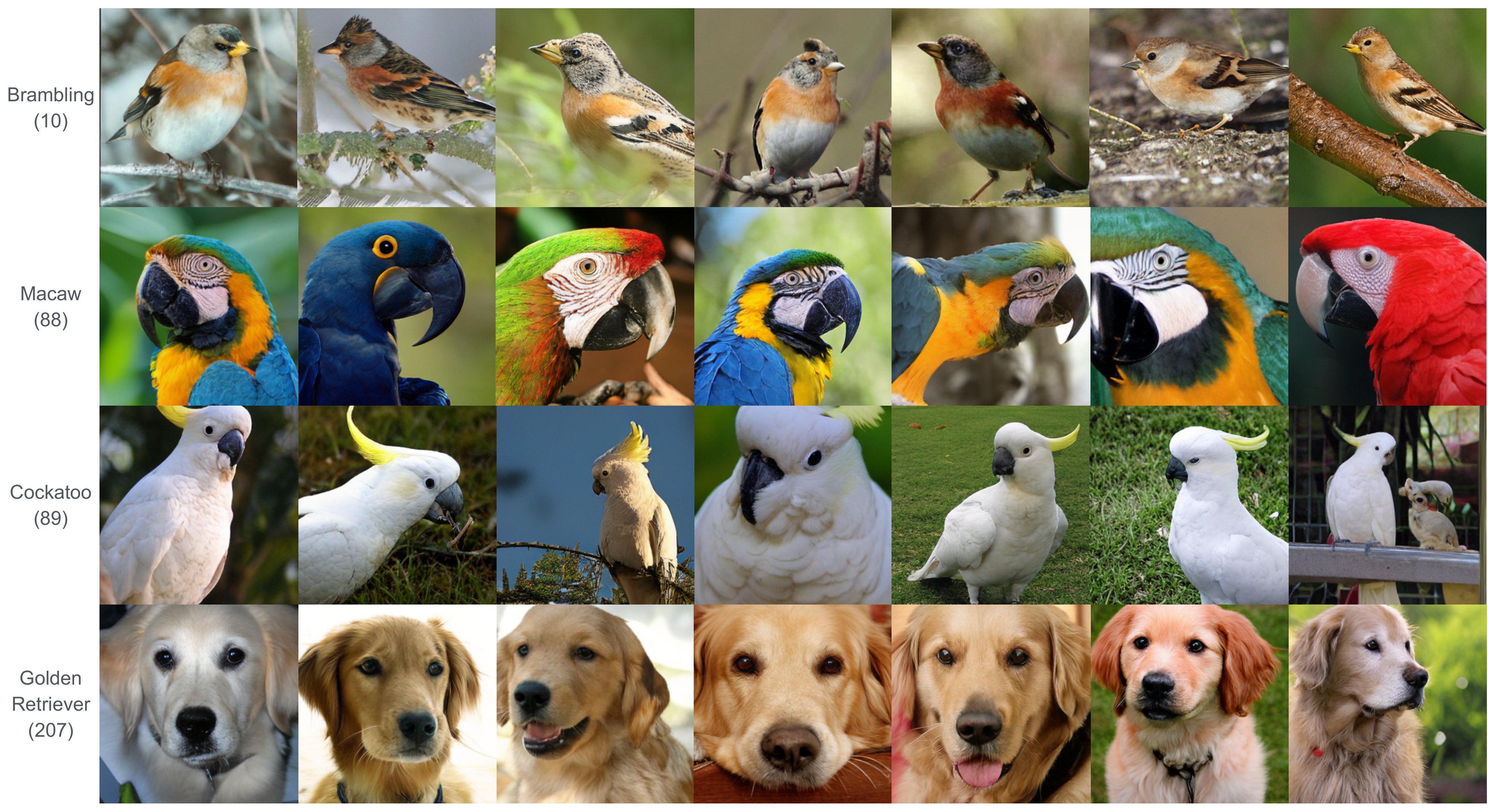}
\end{center}
\caption{\small
\textbf{$256\times 256$ \model{}-XL~(FID=2.07) samples.} Classifier-free guidance scale=1.65. 
}
\label{fig:gen-XL-1}
\end{figure}

\begin{figure}[htb]
\begin{center}
    \includegraphics[width=1.0\textwidth]{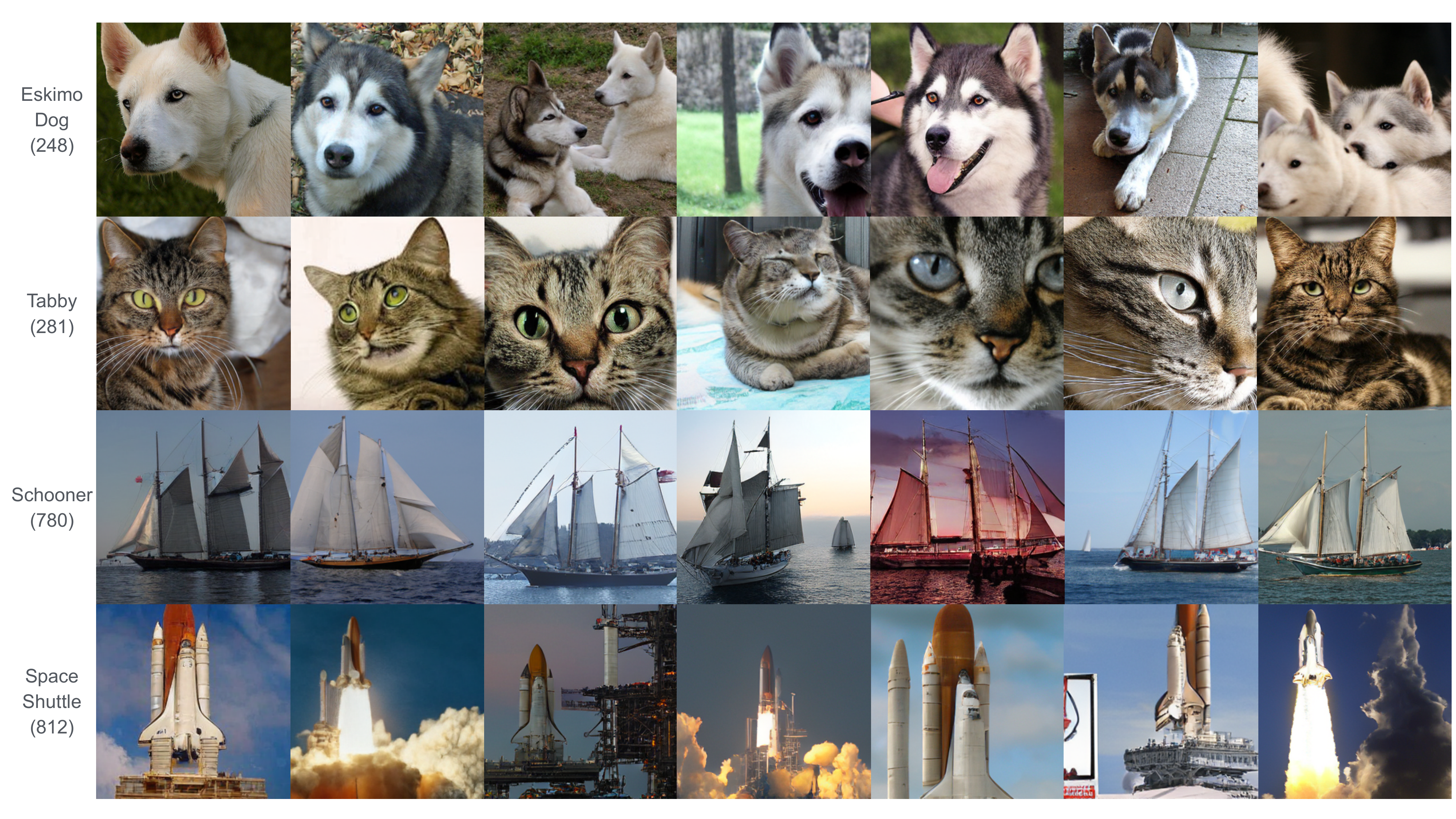}
\end{center}
\caption{\small
\textbf{$256\times 256$ \model{}-XL~(FID=2.07) samples.} Classifier-free guidance scale=1.65. 
}
\label{fig:gen-XL-2}
\end{figure}

\begin{figure}[htb]
\begin{center}
    \includegraphics[width=1.0\textwidth]{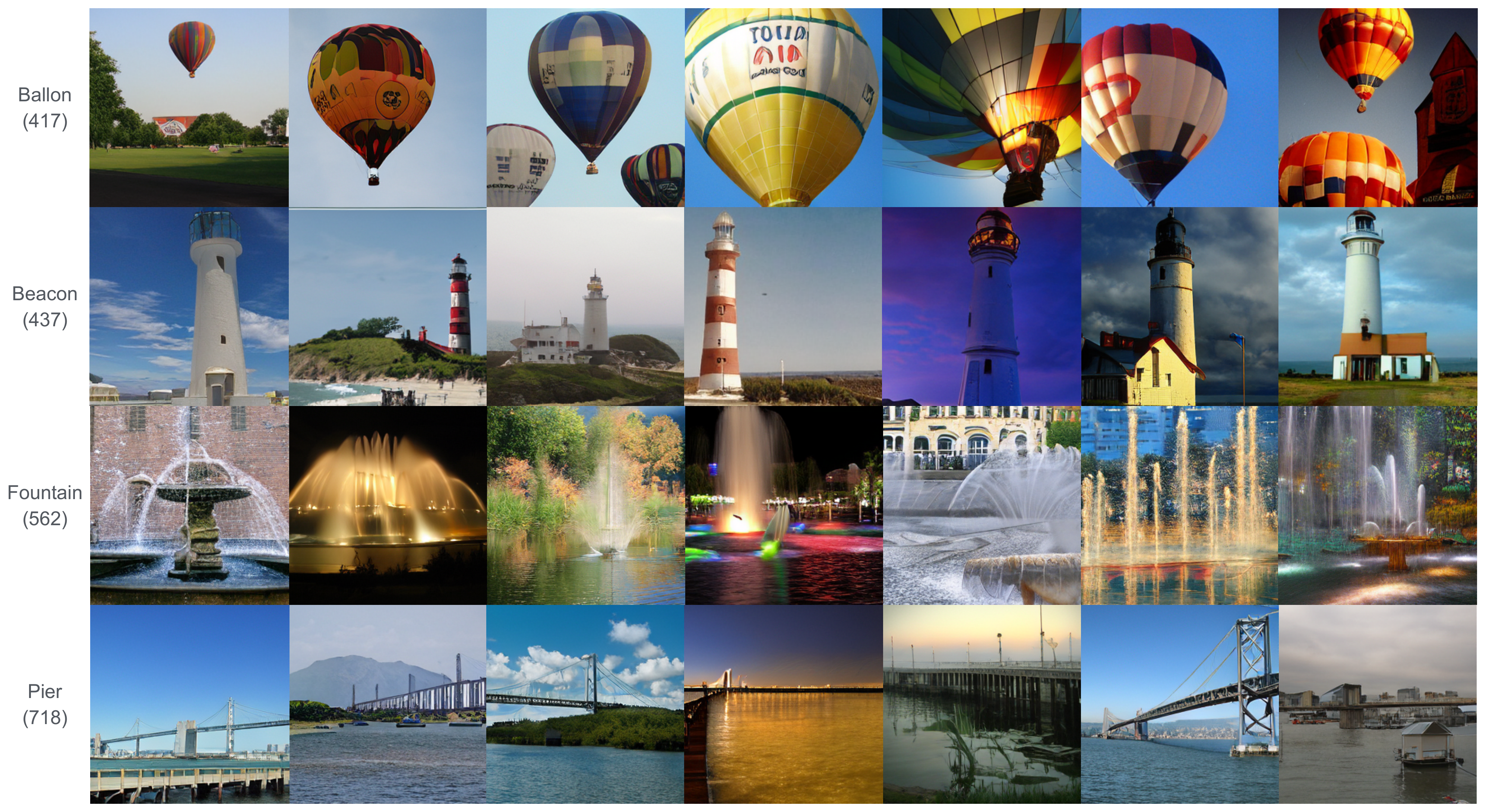}
\end{center}
\caption{\small
\textbf{$256\times 256$ \model{}-XL~(FID=2.07) samples.} Classifier-free guidance scale=1.65. 
}
\label{fig:gen-XL-3}
\end{figure}

\paragraph{Qualitative Reconstruction and Generation Samples.}
We present reconstruction samples in Figure~\ref{fig:recon-demo} and generation samples in Figure~\ref{fig:gen-XL-1}-Figure~\ref{fig:gen-XL-3}.

\end{document}